\documentclass[10pt,twocolumn,letterpaper]{article}

\usepackage[pagenumbers]{cvpr} 
\usepackage{tabularray}

\usepackage{amsmath,amsfonts,bm}

\def\eqref#1{equation~\ref{#1}}

\def\1{\bm{1}}

\def\vx{{\bm{x}}}

\def\vz{{\bm{z}}}

\DeclareMathAlphabet{\mathsfit}{\encodingdefault}{\sfdefault}{m}{sl}
\SetMathAlphabet{\mathsfit}{bold}{\encodingdefault}{\sfdefault}{bx}{n}

\def\gC{{\mathcal{C}}}
\def\gD{{\mathcal{D}}}

\def\gL{{\mathcal{L}}}
\def\gM{{\mathcal{M}}}

\def\gY{{\mathcal{Y}}}

\newcommand{\KL}{D_{\mathrm{KL}}}

\definecolor{cvprblue}{rgb}{0.21,0.49,0.74}
\usepackage[pagebackref,breaklinks,colorlinks,allcolors=cvprblue]{hyperref}

\def\paperID{9943} 
\def\confName{CVPR}
\def\confYear{2025}

\title{Task-Agnostic Guided Feature Expansion for Class-Incremental Learning}

\author{Bowen Zheng\quad Da-Wei Zhou\footnotemark[2]\quad Han-Jia Ye\quad De-Chuan Zhan\\
National Key Laboratory for Novel Software Technology, Nanjing University, China\\
School of Artificial Intelligence, Nanjing University, China\\
{\tt\small \{zhengbw,zhoudw,yehj,zhandc\}@lamda.nju.edu.cn}
}

\begin{document}
\maketitle
\footnotetext[2]{Corresponding author.}
\begin{abstract}
The ability to learn new concepts while preserve the learned knowledge is desirable for learning systems in Class-Incremental Learning (CIL).
Recently, feature expansion of the model become a prevalent solution for CIL, where the old features are fixed during the training of the new task while new features are expanded for the new tasks.
However, such task-specific features learned from the new task may collide with the old features, leading to misclassification between tasks.
Therefore, the expanded model is often encouraged to capture diverse features from the new task, aiming to avoid such collision.
However, the existing solution is largely restricted to the samples from the current task, because of the poor accessibility to previous samples.
To promote the learning and transferring of diverse features across tasks, we propose a framework called \textbf{T}ask-\textbf{A}gnostic \textbf{G}uided \textbf{F}eature \textbf{Ex}pansion (TagFex).
Firstly, it captures task-agnostic features continually with a separate model, providing extra task-agnostic features for subsequent tasks.
Secondly, to obtain useful features from the task-agnostic model for the current task, it aggregates the task-agnostic features with the task-specific feature using a merge attention.
Then the aggregated feature is transferred back into the task-specific feature for inference, helping the task-specific model capture diverse features.
Extensive experiments show the effectiveness and superiority of TagFex on various CIL settings.
Code is available at \url{https://github.com/bwnzheng/TagFex_CVPR2025}.
\end{abstract}    

\section{Introduction}
\label{sec:introduction}

For real-world applications of learning systems, new concepts and knowledge increase over time.
Therefore, the ability to learn new concepts while preserve the learned knowledge is desirable for the learning system.
This motivates the research in Continual Learning (or Incremental Learning).
\emph{Class-Incremental Learning} (CIL)~\cite{masana2022class, zhou2024class,wang2024comprehensive} is one of the scenarios where new concepts incrementally emerge as new classes.
The main challenge for Continual Learning is \emph{catastrophic forgetting}~\cite{mccloskey1989catastrophic}, where the learning system usually forgets previously learned knowledge fast and catastrophically.
Solving catastrophic forgetting alone is not enough for the model to learn new knowledge.
Therefore, researchers seek for balance between stability (ability to resist changes) and plasticity (ability to adapt)~\cite{grossberg2013adaptive}.

\begin{figure}
  \includegraphics[width=\columnwidth]{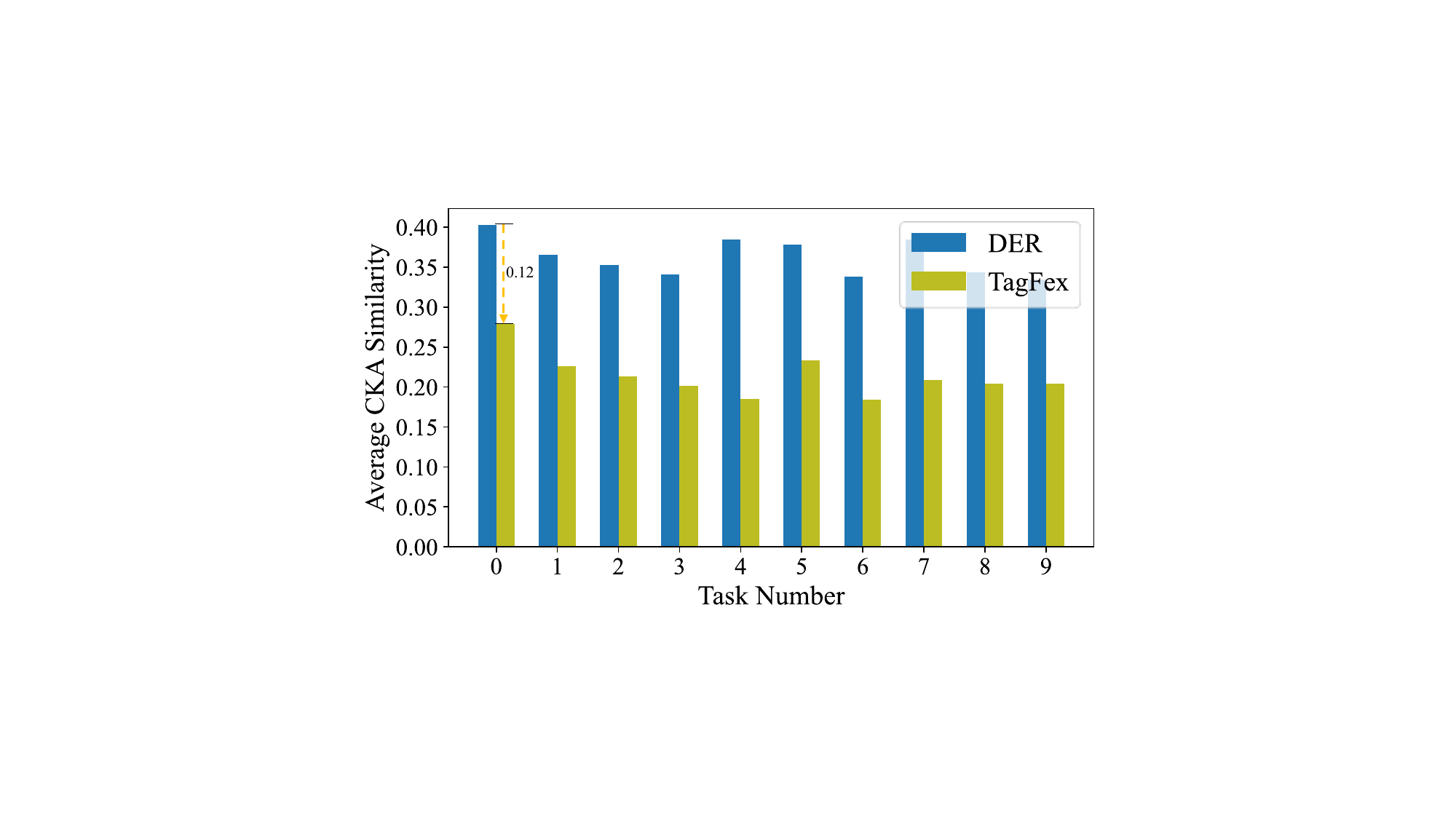}
  \vspace{-3ex}
  \caption{Average CKA feature similarities between different expanded models of DER and TagFex by task. It shows the expanded features learned by TagFex are less correlated.}
  \label{fig:cka}
  \vspace{-2ex}
\end{figure}
\begin{figure}
  \includegraphics[width=\columnwidth]{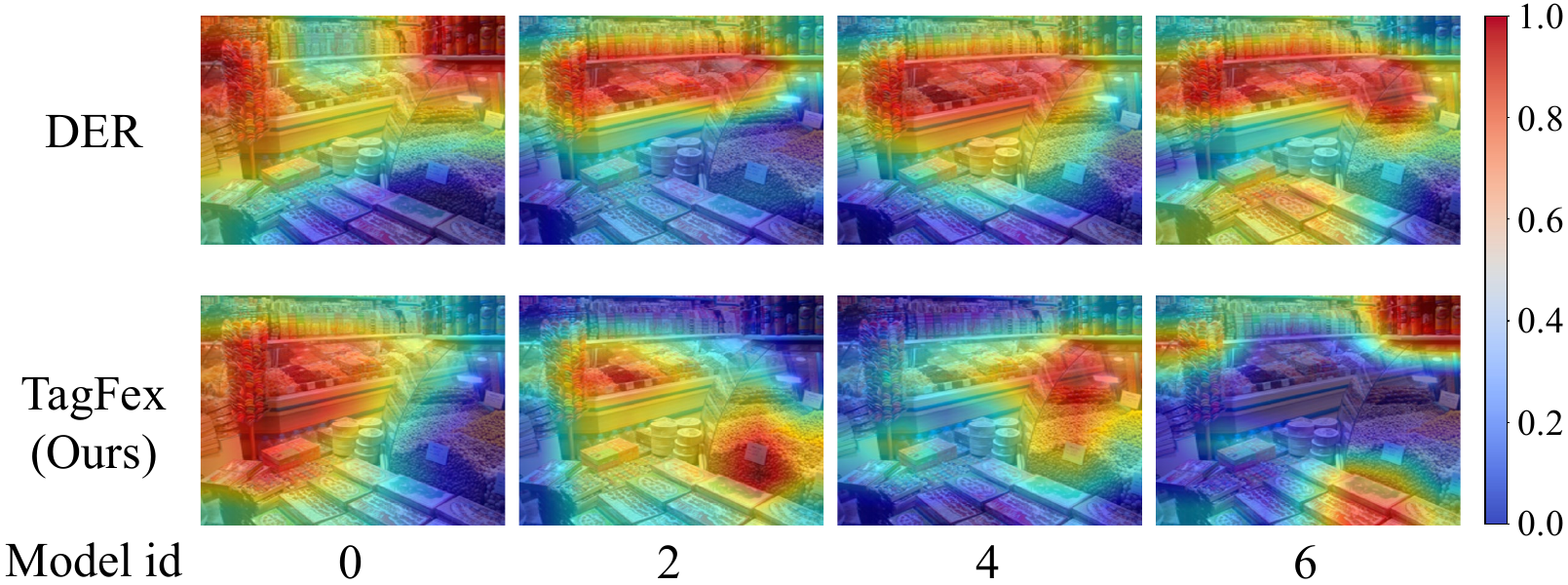}
  \caption{GradCAM visualization comparison between different expanded models by DER and TagFex.
  It shows that the expanded models learned by the proposed framework focus on more diverse features.}
  \label{fig:gradcam}
  \vspace{-4ex}
\end{figure}


\begin{figure*}[!t]
  \vspace{-4ex}
  \begin{center}
    \begin{minipage}{0.35\textwidth}
      \centering
      \includegraphics[width=0.857\textwidth]{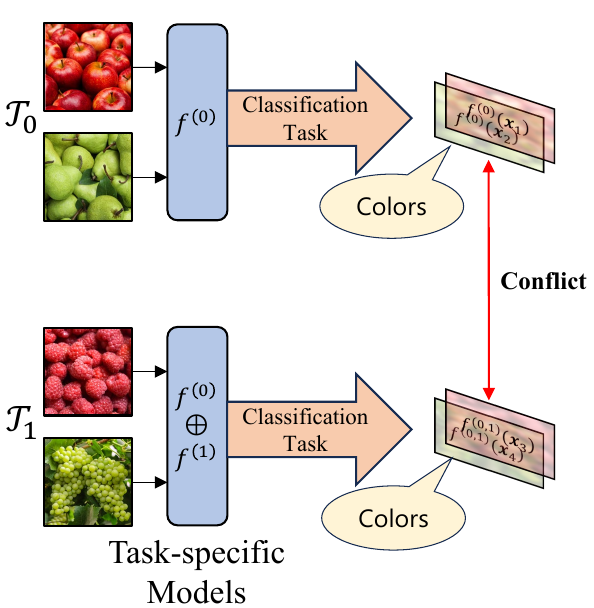}
      \subcaption{Feature Expansion
      Without Rehearsal Samples}
      \label{fig:raw_expansion}
    \end{minipage}
    \quad
    \vrule width 1pt
    \quad
    \begin{minipage}{0.45\textwidth}
      \includegraphics[width=\textwidth]{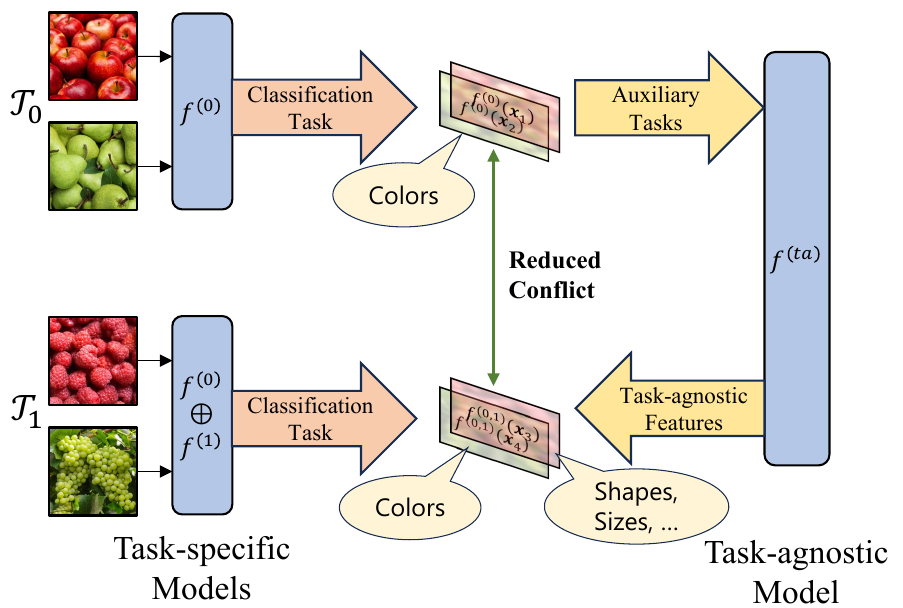}
      \subcaption{Feature Expansion with
      More Diverse Features}
    \end{minipage}
  \end{center}
  \vspace{-3ex}
  \caption{The illustration and mitigation of feature collision in expansion-based CIL. (a) Without the rehearsal samples, it is very difficult to distinguish two types of the fruits with the same color, because the features captured by just training on the newest task are insufficient to distinguish classes between two tasks. (b) With more diverse features, the model is able to further disengage from the spurious features, reducing the feature collision.}
  \label{fig:feature_collision}
  \vspace{-2ex}
\end{figure*}

One of the promising strategies is to train a new feature extractor for each new task while keep the previously learned model intact during the training of the new task.
Such methods are often referred to as \emph{expansion-based} approaches~\cite{yan2021dynamically,wang2022foster,douillard2022dytox,zhou2023model,huang2023resolving}.
Although expansion-based CIL achieve impressive performance on various scenarios in CIL, it relies on the feature diversity extracted by the new model.
Generally, the models trained on different tasks capture different features.
However, there could be new features in the new task colliding with the old ones, leading to classification bias towards the new task.
To understand this, we could consider an extreme case where there are two incremental tasks with similar spurious features, e.g., task 1 with red apples and green pears, task 2 with red berries and green grapes.
It is very difficult to distinguish two types of the fruits with the same color in such incremental scenarios.
As shown in Figure~\ref{fig:raw_expansion}, even if we expand the feature, due to the lack of extra samples for feature capturing, the model would rely on the learned color feature to reach low loss on the new task, and finally make random guesses between the apple and the berry with a sample of red fruits, or simply leaning towards the newest task.

Therefore, to mitigate feature collision, expansion-based methods often introduce an auxiliary classifier and save a small number of rehearsal samples from the previous tasks to distinguish current task samples with previous samples~\cite{yan2021dynamically,douillard2022dytox}.
With the auxiliary classifier, the new model is encouraged to capture diverse features other than the features extracted by the old models.
However, the limited number of rehearsal samples makes the training of the auxiliary classifier imbalanced and suboptimal.
To illustrate this, we perform experiments on the similarities between the learned expanded features.
Specifically, we use CKA~\cite{kornblith2019similarity} as the similarity metric and calculate the similarities between each expanded feature on the initial task.
The average results are shown in Figure~\ref{fig:cka}, and the detailed numbers are shown in Figure~\ref{fig:cka_number} in the Appendix~\ref{app:cka}. 
From the figures, we can conclude that the features extracted by the models learned by existing expansion-based methods are relatively similar, with average CKA around 0.35, especially when comparing to TagFex proposed in this paper, with average CKA around 0.2.
This can also be seen by GradCAM~\cite{selvaraju2017grad, jacobgilpytorchcam} visualizations in Figure~\ref{fig:gradcam}.
We can clearly see that the models learned by DER~\cite{yan2021dynamically} focus on similar areas and those by TagFex focus more diverse areas.

To further encourage the diverse feature capture, a trivial solution would be simply increasing the number of rehearsal samples.
But can we find more efficient approach to help the model acquire more discriminative features?
To this end, we have to help the model to acquire such features that the model has not been able to extract in the classification task and take the advantage of such features for subsequent training.
Figure~\ref{fig:feature_collision} illustrates the general idea and motivation of the proposed framework.

In this paper, we propose to continually capture diverse features in a separate feature space, and then utilize it to help the incremental learning. 
Specifically, for the continual capture of diverse features, we have to find an approach to capture the features agnostic to the classification, which are referred to as \emph{task-agnostic features}.
Therefore, we apply continual self-supervised learning (CSSL)~\cite{guo2022online,fini2022self,cheng2023contrastive} to train a separate task-agnostic model to capture the task-agnostic features continually.
In classification, the models are merely required to capture the minimal necessary features (task-specific features) for the classification within the task.
Whereas, task-agnostic features contain more information that they are hard to be captured in the merely training of the classification task.
Such information is vital for the new model to capture discriminative features from the new task.
To make full utilization of the task-agnostic model, we further design a merge attention to adaptively unify task-agnostic features with task-specific features.
Then, we also transfer the information in the unified representations back into the task-specific model.
Therefore, we are able to mitigate the feature collision in CIL, improving the performance on various scenarios.
The overall framework is called \textbf{T}ask-\textbf{A}gnostic \textbf{G}uided \textbf{F}eature \textbf{Ex}pansion (TagFex).
Moreover, extensive experiments are performed to show the effectiveness of the task-agnostic guided feature expansion and the superiority of TagFex against the state-of-the-art methods, even with memory-aligned protocol.
\section{Related Works}
\label{sec:related_works}
\textbf{Class-Incremental Learning} (CIL) is one of the scenarios in continual learning where the model is learned task by task with a different set of classes.
During inference, the task from which each sample is sampled is not available. 
Many techniques and frameworks are proposed to alleviate catastrophic forgetting and improve the performance in CIL.
They can be roughly categorized into several forms, such as model expansion, rehearsal memory, model distillation and regularization.

\textbf{Model expansion} comes from the idea of parameter isolation for each task.
It expands the feature space for each task.
PNN~\cite{rusu2016progressive} proposes learning a new backbone for each new task and fixing the former in incremental learning.
It also adds layer-wise connections between old and new models to reuse former features.
P\&C~\cite{schwarz2018progress} suggests a progression-compression protocol.
It first expands the network to learn representative representations.
After that, a compression process is conducted to control the total budget.
DER~\cite{yan2021dynamically} trains a separate backbone for each task, aggregating all of the features for classification.
DyTox~\cite{douillard2022dytox} learns a separate task token for each task.
BEEF~\cite{wang2023beef} incorporates energy-based theory to construct a bi-compatible framework.
TCIL~\cite{huang2023resolving} aims to tackle the task confusion, establishing a multi-level knowledge distillation to propagate knowledge learned from the old tasks to the new one.

There are works introducing contrastive learning to continual learning, but from different perspectives.
OCM~\cite{guo2022online} uses contrastive learning to get holistic representations in online incremental scenarios.
C4IL~\cite{ni2021revisiting} introduces contrastive learning on the learned features to alleviate inter-phase confusion in CIL.
Most of them directly apply contrastive learning to the inference model, where the contrastive and classification loss are competing with each other over one model, differing from our proposed TagFex.
We discuss other related works about CIL in Appendix~\ref{app:rel_works}.

\textbf{Continual Self-Supervised Learning} (CSSL) aims to continually learn good representations from non-stationary data.
Rehearsal-based method LUMP~\cite{madaan2022representational} utilizes rehearsal samples to augment current task samples by mixup.
Regularization-based method CaSSLe~\cite{fini2022self} encourages current model to maintain a consistency with previous state via a prediction head.
C$^2$ASR~\cite{cheng2023contrastive} selects the most representative and discriminative samples by estimating the augmentation stability for rehearsal.

In this paper, we treat the features in such representations as task-agnostic features, since they are agnostic to the classification task, which contain the information that task-specific features do not contain.
To the best of our knowledge, using task-agnostic features to mitigate feature collision in expansion-based CIL has rarely been explored.
Our work propose to continually capture task-agnostic features in each task, and utilizes such features to help the expanded feature acquire more diverse features, alleviating feature collision.
\section{Preliminaries}
\subsection{Problem Formulation}
In CIL scenarios, we have multiple classification tasks to learn sequentially. 
Let $\gD_t$ be the training dataset of the $t$th task. 
$(\vx_i^{(t)}, y_i^{(t)})\in \gD_t$ is a sample. 
$\vx_i^{(t)}$ is the input, $y_i^{(t)}$ is the label. 
Let $\gC_t=\bigcup_i\{y_i^{(t)}\}$ be the class set of task $t$.
In CIL, $\forall t_1\neq t_2, \gC_{t_1}\cap\gC_{t_2}=\emptyset$. 
In each task, we only train the model on $\gD_t$, but test on all the tasks the model has trained on, i.e., the seen tasks. 
For example, when the model is training on task $t_i$, the seen tasks are tasks $t_j (j\le i)$.
Therefore, the class set of the seen tasks is $\gY_t = \bigcup_t\gC_t$.
In rehearsal-based CIL, a small number of rehearsal samples $\gM$ are allowed to be stored for later tasks.
The number of rehearsal samples should be constant during the CIL training.
The goal is to make the model get better performance on all the seen tasks.

\begin{figure*}[!t]
  \vspace{-4ex}
  \centering
  \includegraphics[width=0.8\textwidth]{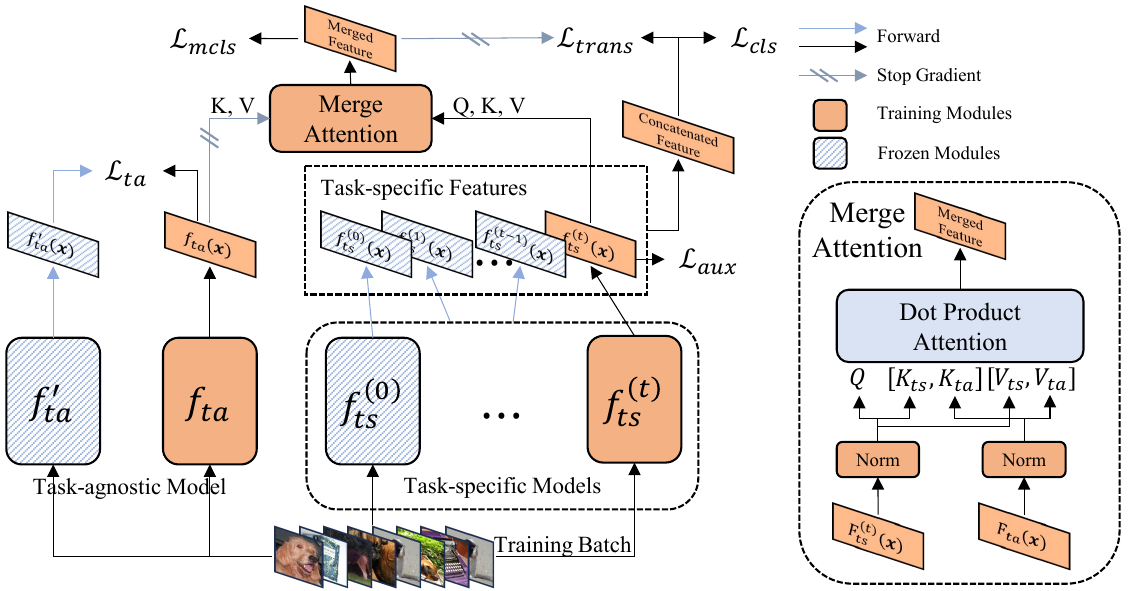}
  \caption{Overview of the proposed framework TagFex.
  The framework is composed of three parts: \emph{Task-agnostic Model} (Section~\ref{sec:acq_ta}), \emph{Merge Attention} (Section~\ref{sec:merge_attn}) and \emph{Task-specific Models} (Section~\ref{sec:ts_expansion} and \ref{sec:ts_trans}).
  Overall, our framework continually captures the task-agnostic feature in each task and transfers such feature back to the task-specific feature, helping the task-specific model obtain more diverse features to combat the feature collision.}
  \label{fig:overview}
  \vspace{-3ex}
\end{figure*}

\subsection{Feature Expansion of Task-Specific Models}
\label{sec:ts_expansion}
In this section we introduce DER~\cite{yan2021dynamically} as the basis of the feature expansion for our task-specific models.
As shown in the middle part of Figure~\ref{fig:overview} (Task-specific Models), a new model $f^{(t)}_{\mathrm{ts}}$ is created at the beginning of each task.
Also, the previously learned models $f^{(0)}_{\mathrm{ts}}\sim f^{(t-1)}_{\mathrm{ts}}$ are saved and freezed for feature extraction.
The classifier is also expanded and the trained parameter weights from the old classifier is inherited.
To predict the label of a sample, the concatenated feature from all of the models is used by the classifier.
Formally, the classification loss $\gL_{\mathrm{cls}}$ with cross-entropy (CE) for the sample $(\vx,y)$ would be,
\begin{equation}
  \gL_{\mathrm{cls}}(\vx, y) = \ell_{\mathrm{CE}}([f_{\mathrm{ts}}^{(0)}(\vx),\dots,f_{\mathrm{ts}}^{(t)}(\vx)], y; W_{\mathrm{cls}}^{(t)}),
\end{equation}
where $y\in\gY_t$ and $W_{\mathrm{cls}}^{(t)}$ is the parameter weights of the classifier in task $t$.

Additionally, to help the new model to capture diverse features, an auxiliary classifier is trained based on the new feature.
The training objective for the auxiliary classifier is to not only distinguish classes within the current task, but also distinguish the samples from the previous tasks.
Consequently, a small number of rehearsal samples are saved from the previous tasks.
Therefore, the auxiliary classifier outputs $|\gC_t|+1$ logits, where the old classes are treated as a whole.
Formally, the auxiliary loss $\gL_{\mathrm{aux}}$ with cross-entropy for the sample $(\vx,y)$ would be,
\begin{equation}
  \gL_{\mathrm{aux}}(\vx, y) = \ell_{\mathrm{CE}}(f_{\mathrm{ts}}^{(t)}(\vx), y; W_{\mathrm{aux}}^{(t)}),
\end{equation}
where $y\in[|\gC_t|+1]$ and $W_{\mathrm{aux}}^{(t)}$ is the parameter weights of the auxiliary classifier in task $t$.

\subsection{Discussions on the Feature Diversity in Task-Specific Models}
In supervised classification within one task, the model is only required to capture the minimal necessary features for the classification within the task.
For example, in binary classification of \emph{rabbit} and \emph{whale}, the model only needs to distinguish the colors of the background and the object to minimize the classification loss.
Such features are useful within a specific task, we thus refer to such features as \emph{task-specific features}.
Therefore, the task-specific models trained by DER can only extract the task-specific features for each task.
However, apart from the task-specific features, there are plenty of features not discovered by these models.
Such features are not necessarily related to the classification of the task, so we refer them as \emph{task-agnostic features}.
If these features are available for subsequent tasks, the new task-specific model can refer to them for better feature learning.

DER uses rehearsal samples and the auxiliary classifier to capture diverse features, which makes the situation better.
However, since the number of rehearsal samples are limited, the imbalanced training makes it suboptimal to capture more diverse features.
Therefore, to further improve the ability to capture more diverse features, we explore how to incorporate the task-agnostic feature into this task-specific features in the next section.
\section{Task-Agnostic Guided Feature Expansion}
  \subsection{Continual Acquisition of Task-Agnostic Features}
  \label{sec:acq_ta}
  Inspired by recent studies from continual self-supervised learning~\cite{fini2022self, madaan2022representational, cheng2023contrastive}, not only can we extract task-agnostic features with self-supervised learning, but also continually extract new task-agnostic features while keeping the old ones.
  For the continual acquisition of task-agnostic features, we follow CaSSLe~\cite{fini2022self} to continually capture more features in a predictive manner.
  Also, we take SimCLR~\cite{chen2020simple} as the base self-supervised method for CaSSLe and discuss other self-supervised methods in section~\ref{sec:further_analysis}.
  As shown in the left part of Figure~\ref{fig:overview} (Task-agnostic Model), we train a separate model to maintain the task-agnostic feature for the learned tasks.
  So that, the task-agnostic feature will not be affected by the classification objectives.
  Also, we save the trained task-agnostic model for the next task, in order to support the predictive knowledge preservation in CaSSLe.
  The overall loss for the task-agnostic model in the initial task and incremental tasks are,
{ \allowdisplaybreaks 
  \begin{align}
    \gL_{\mathrm{ta}}^{(t=0)} =& -\mathrm{InfoNCE}(\{f_{\mathrm{ta}}(\vx_i)\}_{i=1}^B, \{f_{\mathrm{ta}}(\vx_i')\}_{i=1}^B), \label{eq:infonce1}\\
    \gL_{\mathrm{ta}}^{(t>0)} =& -\mathrm{InfoNCE}(\{f_{\mathrm{ta}}(\vx_i)\}_{i=1}^B, \{f_{\mathrm{ta}}(\vx_i')\}_{i=1}^B) \label{eq:infonce2}\\
    &- \mathrm{InfoNCE}(\{g(f_{\mathrm{ta}}(\vx_i))\}_{i=1}^B, \{f_{\mathrm{ta}}'(\vx_i)\}_{i=1}^B), \label{eq:infonce_predict}\\
    \mathrm{where}\quad&\mathrm{InfoNCE}(\{\vz_i\}_{i=1}^B,\{\vz'\}_{i=1}^B) \nonumber\\
    =& \frac{1}{B}\sum_{i=1}^{B}{\log{\frac{\exp(\mathrm{sim}(\vz_i, \vz_i')/\tau)}{\sum_{j=1, j\neq i}^{B}{\exp(\mathrm{sim}(\vz_i,\vz_j')/\tau)}}}}, \label{eq:infonce}
  \end{align}}
  Equation~\ref{eq:infonce1} and \ref{eq:infonce2} are contrastive InfoNCE~\cite{oord2018representation} loss to maximize the lower bound of the mutual information between the sample and its representation.
  The last term in equation~\ref{eq:infonce_predict} aims to train a predictor $g(\cdot)$ to predict the last task-agnostic feature ($f'_{\mathrm{ta}}(\vx_i)$) with the current one ($f_{\mathrm{ta}}(\vx_i)$), where $f'_{\mathrm{ta}}(\cdot)$ is the last task-agnostic model we save.
  $\mathrm{sim}(\cdot,\cdot)$ in Equation~\ref{eq:infonce} includes another projection head where we apply cosine similarity in the projected feature space.
  The core idea of the predictive loss is that if such loss is minimized, the current feature at least contains as much information as the last feature contains.
  Therefore, the expressiveness of the task-agnostic feature is increasing task by task as the incremental learning proceeds.
  
  \subsection{Adaptive Feature Aggregation by Merge Attention}
  \label{sec:merge_attn}
  Now we have a powerful feature extractor to continually capture the task-agnostic features from all of the tasks we have seen.
  To put the task-agnostic model into good use, we design the \emph{merge attention} to adaptively unify task-agnostic features with task-specific features in the incremental tasks.
  
  As shown in the right part of Figure~\ref{fig:overview} (Merge Attention), the merge attention takes the feature maps of both the task-agnostic and task-specific features of a sample as inputs, denoting them as $F_{\mathrm{ta}}(\vx)$ and $F_{\mathrm{ts}}(\vx)$ respectively.
  Like the patch embedding of ViT~\cite{dosovitskiy2021an}, we treat each spacial feature in the feature map ($H\times W$ features) as a patch token in our attention.
  The feature maps are first normalized by two layer norms for training stability.
  \begin{align}
    \vz_{\mathrm{ts}} &= \mathrm{LayerNorm_1}(F_{\mathrm{ts}}(\vx)), \\
    \vz_{\mathrm{ta}} &= \mathrm{LayerNorm_2}(\mathrm{StopGradient}(F_{\mathrm{ta}}(\vx))).
  \end{align}
  Then, in order to absorb useful features from task-agnostic features, we take $\vz_{\mathrm{ts}}$ as the query, making them attend to both $\vz_{\mathrm{ts}}$ and $\vz_{\mathrm{ta}}$.
  Furthermore, since $\vz_{\mathrm{ts}}$ and $\vz_{\mathrm{ta}}$ are in different feature spaces, we calculate their keys and values separately.
  \begin{eqnarray}
    Q = W_q\vz_{\mathrm{ts}}, &K_{\mathrm{ts}} = W_{k}^{(\mathrm{ts})}\vz_{\mathrm{ts}}, & V_{\mathrm{ts}} = W_{v}^{(\mathrm{ts})}\vz_{\mathrm{ts}}, \\
    &K_{\mathrm{ta}} = W_{k}^{(\mathrm{ta})}\vz_{\mathrm{ta}}, & V_{\mathrm{ta}} = W_{v}^{(\mathrm{ta})}\vz_{\mathrm{ta}}.
  \end{eqnarray}
  We then concatenate their keys for the calculation of the attention map.
  Finally, we get the merged feature map by multi-head scaled dot product attention~\cite{bahdanau2014neural, vaswani2017attention} with concatenated keys and values.
  For each head of the embedding:
  \begin{eqnarray}
    O^{(h)} = \mathrm{Softmax}(\frac{Q^{(h)}[K_{\mathrm{ts}}^{(h)}, K_{\mathrm{ta}}^{(h)}]^T}{\sqrt{d/h}})[V_{\mathrm{ts}}^{(h)}, V_{\mathrm{ta}}^{(h)}],
  \end{eqnarray}
  where $d$ is the embedding dimension of each patch token, $h$ is the number of heads.
  Such extended self-attention makes the queries not only attend to task-specific features themselves, but also attend to task-agnostic features.
  Therefore, the merged feature contains not only the information learned from the current task, but also relevant information from the task-agnostic feature.
  
  To guide the merge process, we use global average pooling to get the feature for the merge classifier.
  The merge classifier is trained to predict the label of the current task samples.
  \begin{equation}
    \gL_{\mathrm{mcls}} = \ell_{\mathrm{CE}}(\mathrm{GAP}(O), y).
  \end{equation}
  Note that we do not propagate the gradients to the task-agnostic model.
  Therefore, no task-specific information about the classification would affect the task-agnostic model, keeping the task-agnostic feature uncorrelated with task-specific ones.

\begin{table*}[!t]
    \vspace{-4ex}
    \centering
    \caption{Performance results. Bold font represents the best results in the scenario. TagFex-P is the method with pruning strategy.}
    \label{tab:performance}
    \scalebox{1.}{
    \begin{tblr}{
      rows = {abovesep=1.5pt, belowsep=1.5pt},
      row{1} = {c},
      row{2} = {c},
      row{3} = {c},
      column{3,7} = {rightsep=15pt},
      column{5,9} = {rightsep=20pt},
      column{2-11} = {c},
      cell{1}{1} = {r=3}{},
      cell{1}{2} = {c=4}{},
      cell{1}{6} = {c=4}{},
      cell{1}{10} = {c=2}{},
      cell{2}{2} = {c=2}{},
      cell{2}{4} = {c=2}{},
      cell{2}{6} = {c=2}{},
      cell{2}{8} = {c=2}{},
      cell{2}{10} = {c=2}{},
      hline{1,11} = {-}{0.08em},
      hline{2-3} = {2-11}{},
      hline{4,9} = {-}{},
    }
    Methods                         & CIFAR100       &                &                &                & ImageNet100    &                &                &                & ImageNet1000   &                \\
    & 10-10          &                & 50-10          &                & 10-10          &                & 50-10          &                & 100-100        &                \\
    & Last           & Avg            & Last           & Avg            & Last           & Avg            & Last           & Avg            & Last           & Avg            \\
iCaRL~\cite{rebuffi2017icarl}   & 49.52          & 64.64          & 50.56          & 60.08          & 50.98          & 67.11          & 53.69          & 62.56          & 40.47          & 57.55          \\
BiC~\cite{wu2019large}          & 50.79          & 65.38          & 43.82          & 57.04          & 42.40          & 65.13          & 49.90          & 66.36          & -              & -              \\
DyTox~\cite{douillard2022dytox} & 60.43          & 73.50          & -              & -              & 67.61          & 76.51          & -              & -              & 59.75          & 68.14          \\
BEEF~\cite{wang2023beef}        & 60.98          & 71.94          & 63.51          & 70.71          & 68.78          & 77.62          & 70.98          & 77.27          & 58.67          & 67.09          \\
DER~\cite{yan2021dynamically}   & 64.35          & 75.36          & 65.27          & 72.60          & 66.71          & 77.18          & 71.08          & 77.71          & 58.83          & 66.87          \\
TagFex                          & \textbf{68.23} & \textbf{78.45} & \textbf{70.33} & \textbf{75.87} & \textbf{70.84} & \textbf{79.27} & \textbf{75.54} & \textbf{80.64} & \textbf{61.45} & \textbf{68.32} \\
TagFex-P                        & 67.34          & 78.02          & 69.26          & 74.24          & 69.21          & 78.56          & 74.13          & 79.85          & 60.14          & 67.65          
\end{tblr}}
\end{table*}
  \begin{figure*}
      \begin{minipage}{0.245\textwidth}
          \includegraphics[width=\textwidth]{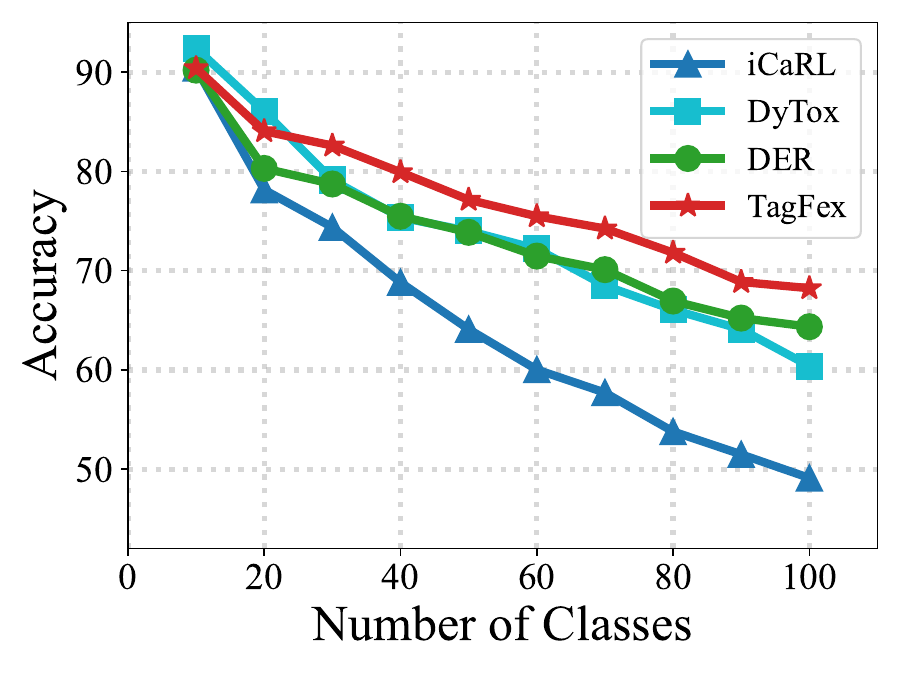}
          \subcaption{CIFAR100 10-10}
      \end{minipage}
      \hfill
      \begin{minipage}{0.245\textwidth}
          \includegraphics[width=\textwidth]{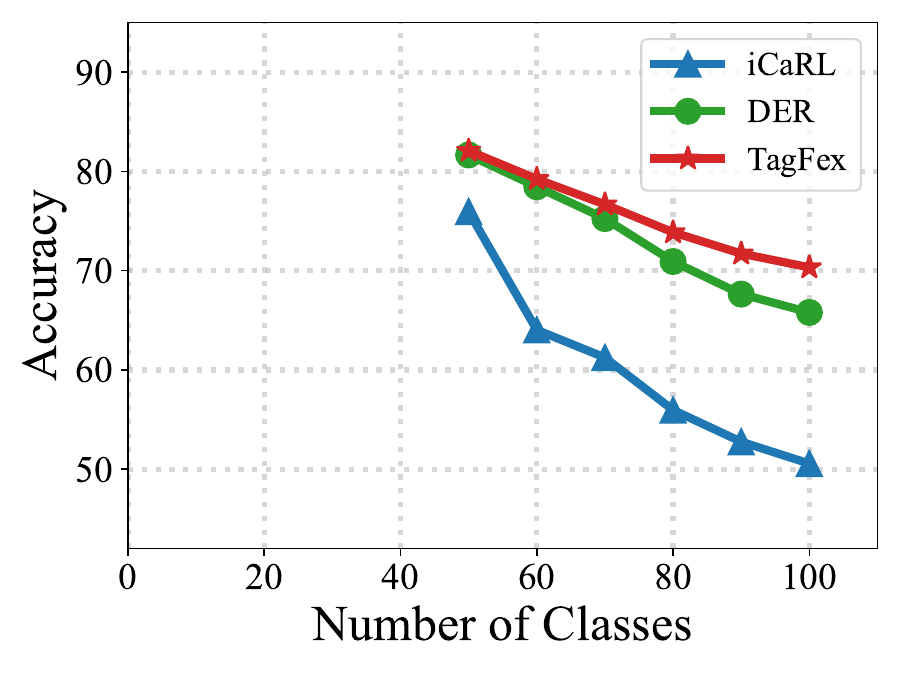}
          \subcaption{CIFAR100 50-10}
      \end{minipage}
      \hfill
      \begin{minipage}{0.245\textwidth}
          \includegraphics[width=\textwidth]{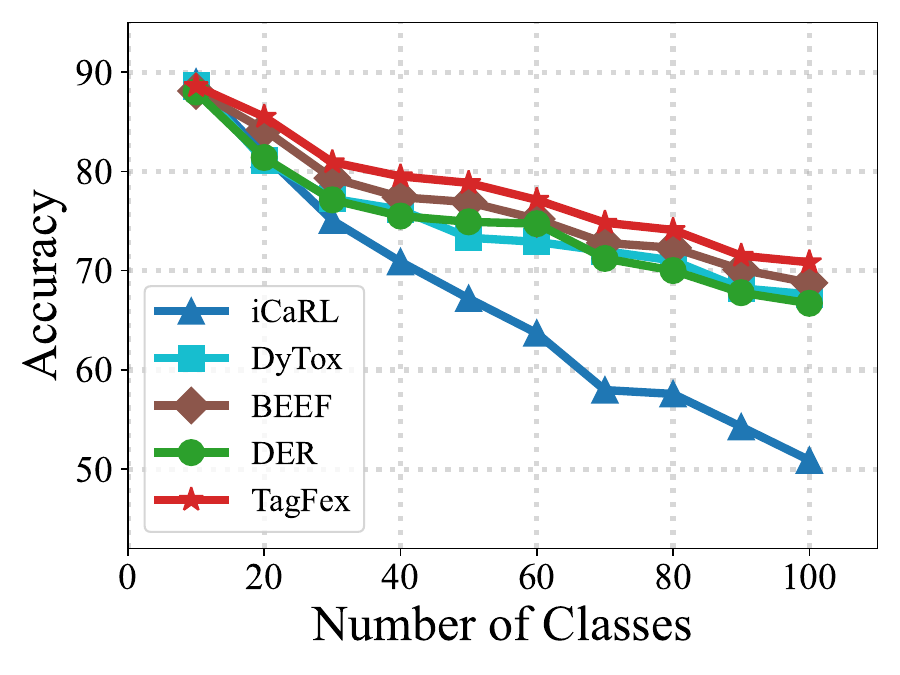}
          \subcaption{ImageNet100 10-10}
      \end{minipage}
      \hfill
      \begin{minipage}{0.245\textwidth}
          \includegraphics[width=\textwidth]{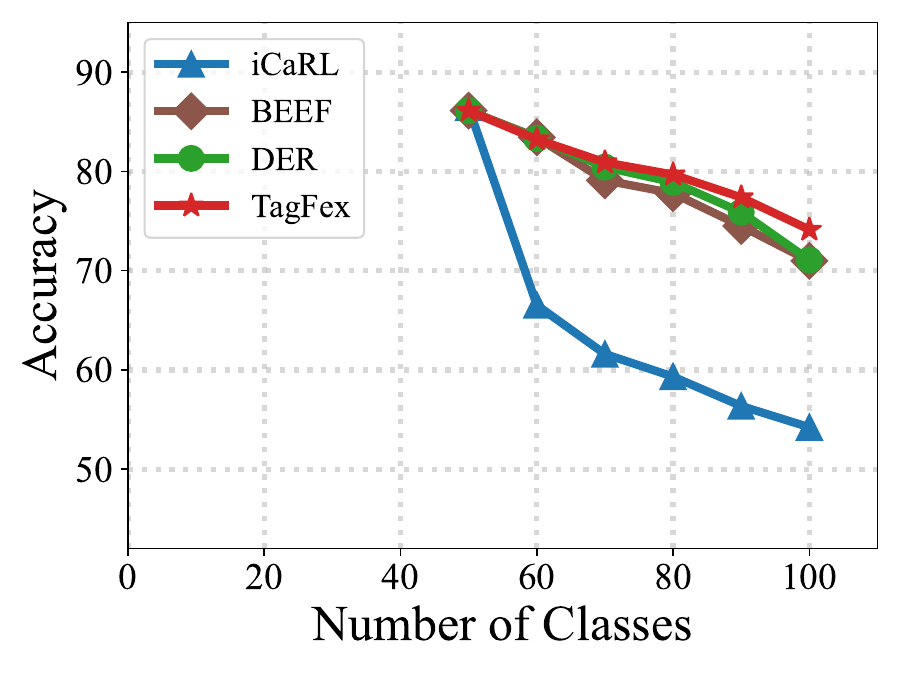}
          \subcaption{ImageNet100 50-10}
      \end{minipage}
      \vspace{-1ex}
      \caption{Accuracy curves for baselines and TagFex on various scenarios.}
      \label{fig:acc_curve}
      \vspace{-2ex}
  \end{figure*}

  \subsection{Task-agnostic Guided Knowledge Transferring}
  \label{sec:ts_trans}
  With the merge attention, we have obtained more diverse features and their corresponding logits.
  However, we do not directly use the logits of the merge classifier for the final inference, since the distributional drift happens when the task-agnostic model is continually updated as the incremental training goes on.
  We have to transfer the diverse features from merged features to task-specific features.
  In the end, we will only use the task-specific models with the classifier for inference.
  
  For the knowledge transferring, we use KL divergence between the logits of the merge classifier ($\log(p_{\mathrm{m}})$) and task-specific classifier ($\log(p_{\mathrm{ts}})$) whose weights $W_{\mathrm{cls}}^{(t)}$ are mentioned in section~\ref{sec:ts_expansion} within the current task, which is,
  \begin{equation}
    \gL_{\mathrm{trans}} = {\KL(\mathrm{StopGradient}(p_{\mathrm{m}}^{(i)})||p_{\mathrm{ts}}^{(i)})}.
  \end{equation}
  Therefore, with the losses mentioned in section~\ref{sec:ts_expansion}, the overall loss we apply on the task-specific model is,
  \begin{equation}
    \gL_{\mathrm{ts}} = \gL_{\mathrm{cls}} + \gL_{\mathrm{aux}} + \gL_{\mathrm{trans}}.
  \end{equation}
  
  In addition, since the expansion of the task-specific models suffers from rapidly increasing number of parameters at both training and inference.
  To eliminate such drawback for practical use of our method, we adopt the FPGM~\cite{he2019filter} as the pruning strategy.
  It takes the geometric median within the kernel pairs as the similarity.
  Then the kernel pairs whose similarity above a threshold are treated as redundant and one of them is pruned.
  With this strategy, the model size could be largely reduced while not affecting the accuracy too much.
  Since TagFex only uses task-specific models for inference, we only prune the task-specific models.
  With such pruning strategy, the method is called \emph{TagFex-P}.

  \subsection{Summary}
  In summary, the proposed framework TagFex is composed of three parts: 1) the continual acquisition of the task-agnostic features, 2) the adaptive feature aggregation with merge attention, 3) the task-agnostic guided knowledge transferring.
  The task-agnostic features are acquired continually by a separate model, and such task-agnostic information is first aggregated with the task-specific features and then transferred back into the task-specific features.
  The overall loss of our framework during the training of each task is the weighted sum of the three parts,
  \begin{equation}
    \gL = \lambda_{\mathrm{ta}}\gL_{\mathrm{ta}} + \lambda_{\mathrm{mcls}}\gL_{\mathrm{mcls}} + \gL_{\mathrm{ts}}.
  \end{equation}
  For practical use and fair comparison, we further adopt pruning strategy to compress the task-specific models without affecting the accuracy too much.
\section{Experiments}
\subsection{Experimental Setups}
\label{sec:exp_setups}
\textbf{Datasets and Rehearsal Memory}.
Following most of the image classification benchmarks in CIL~\cite{rebuffi2017icarl,wu2019large}, we use CIFAR100 and ImageNet to train the model incrementally.
CIFAR100~\cite{Krizhevsky2009LearningML} has 50,000 training and 10,000 testing samples with 100 classes in total.
Each sample is a tiny image in 32 $\times$ 32 pixels.
ImageNet~\cite{deng2009imagenet} has 1,300 training samples and 50 test samples for each class.
The maximum number of rehearsal samples is 2,000 as default for CIFAR100 and ImageNet100, and 20,000 for ImageNet1000.
We use \emph{herding}~\cite{rebuffi2017icarl} for the selection and eviction of the rehearsal samples between each task.

\textbf{Data Split}.
There are two common types of splits in CIL. 
The \emph{small base} one equally divides all of the classes in a dataset~\cite{rebuffi2017icarl}.
The \emph{large base} one uses half of the classes in a dataset as the base task (task 0), and equally divides the remaining classes~\cite{hou2019learning, yu2020semantic}.
For a dataset with 100 classes, \emph{10-10} means 10 classes in the base task and incremental tasks are also with 10 classes each, \emph{50-10} means 50 classes in the base task and 10 classes in the incremental tasks.

\textbf{Backbones and Implementation}.
We follow DER~\cite{yan2021dynamically} to use ResNet18~\cite{he2016deep} as the backbone model on both CIFAR and ImageNet for both task-agnostic and task-specific models.
The experiments are implemented based on the open-source code PyCIL~\cite{zhou2023pycil}.
We provide more implementation details in Appendix~\ref{app:more_imp}.

\begin{figure*}[!t]
  \vspace{-4ex}
  \begin{minipage}{0.05\textwidth}
    \includegraphics[width=\textwidth]{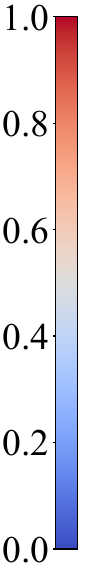}
    \vspace{1ex}
  \end{minipage}
  \begin{minipage}{0.95\textwidth}
    \begin{minipage}{0.32\textwidth}
      \includegraphics[width=\textwidth]{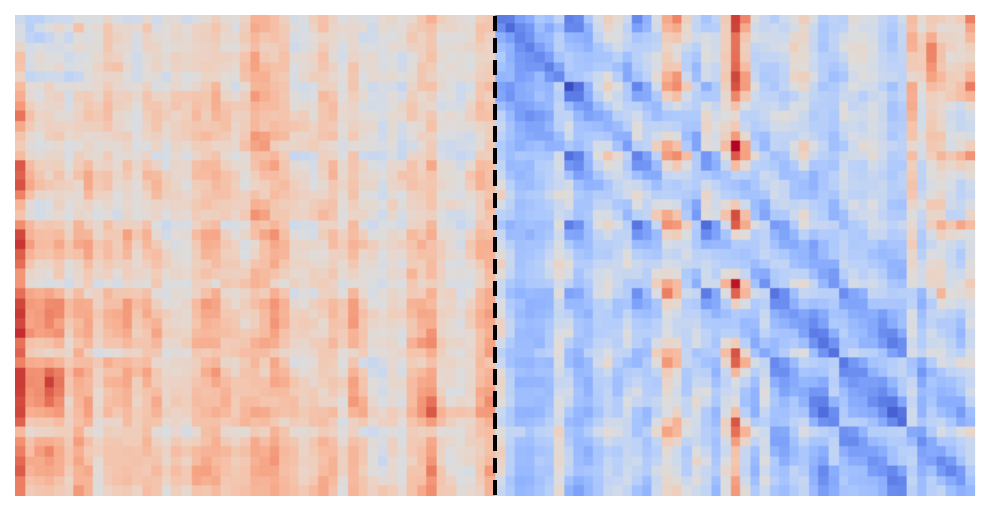}
      \subcaption{Task 2 Epoch 0}
    \end{minipage}
    \hfill
    \begin{minipage}{0.32\textwidth}
      \includegraphics[width=\textwidth]{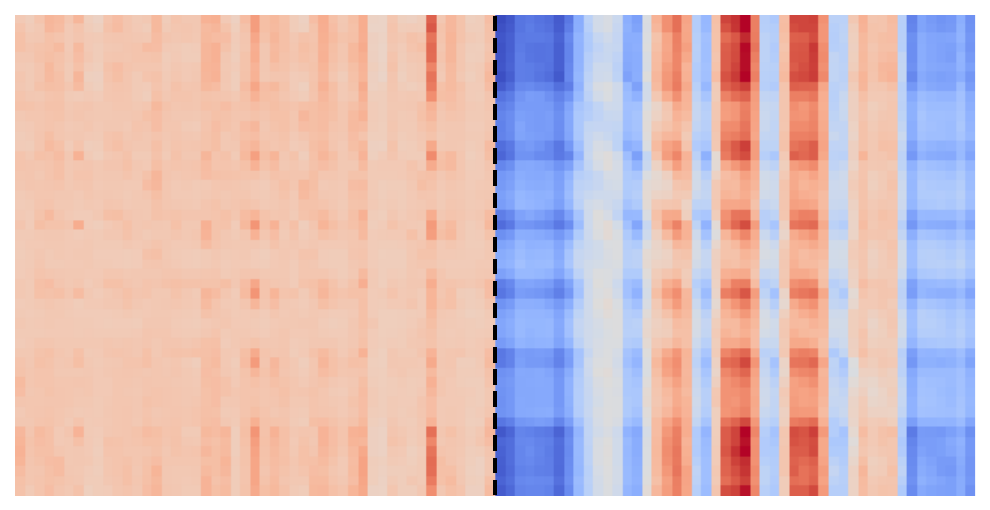}
      \subcaption{Task 2 Epoch 80}
    \end{minipage}
    \hfill
    \begin{minipage}{0.32\textwidth}
      \includegraphics[width=\textwidth]{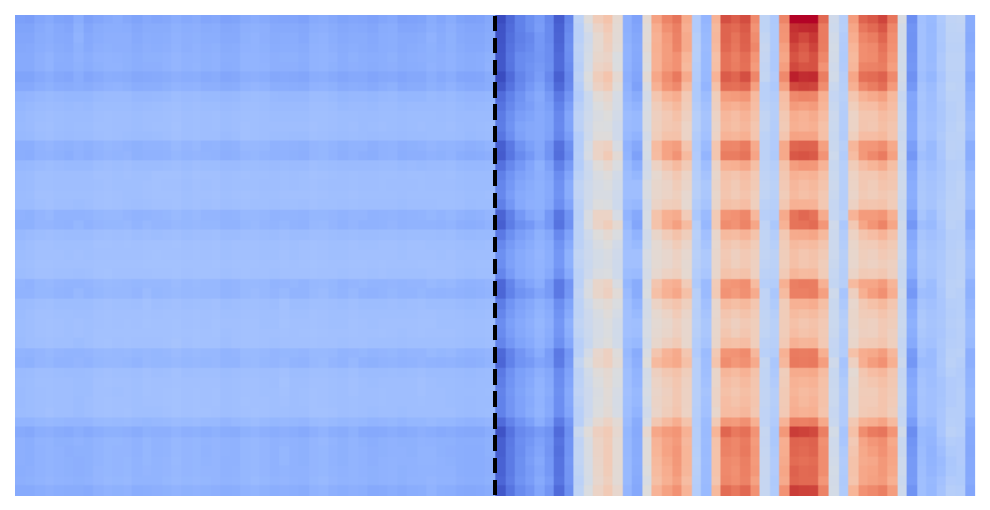}
      \subcaption{Task 2 Epoch 160}
    \end{minipage}
    \begin{minipage}{0.32\textwidth}
      \includegraphics[width=\textwidth]{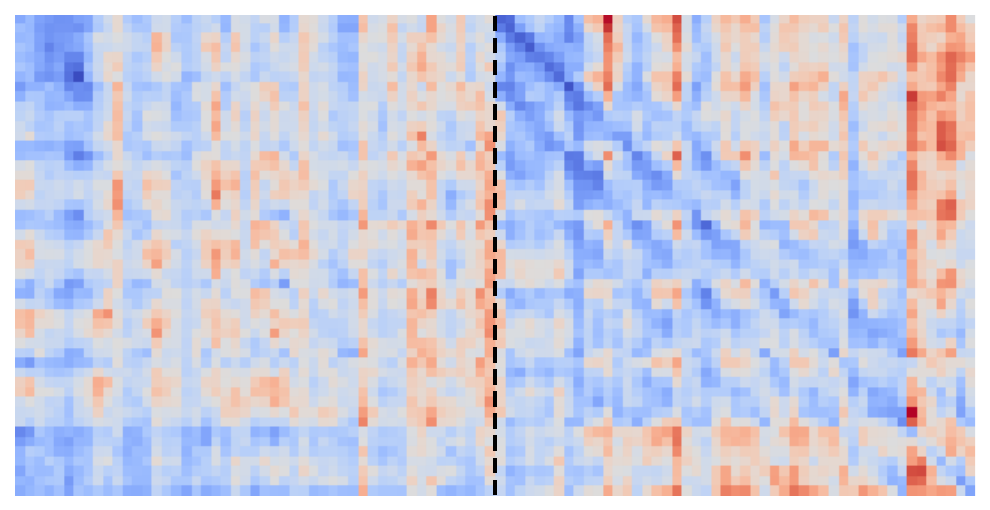}
      \subcaption{Task 4 Epoch 0}
    \end{minipage}
    \hfill
    \begin{minipage}{0.32\textwidth}
      \includegraphics[width=\textwidth]{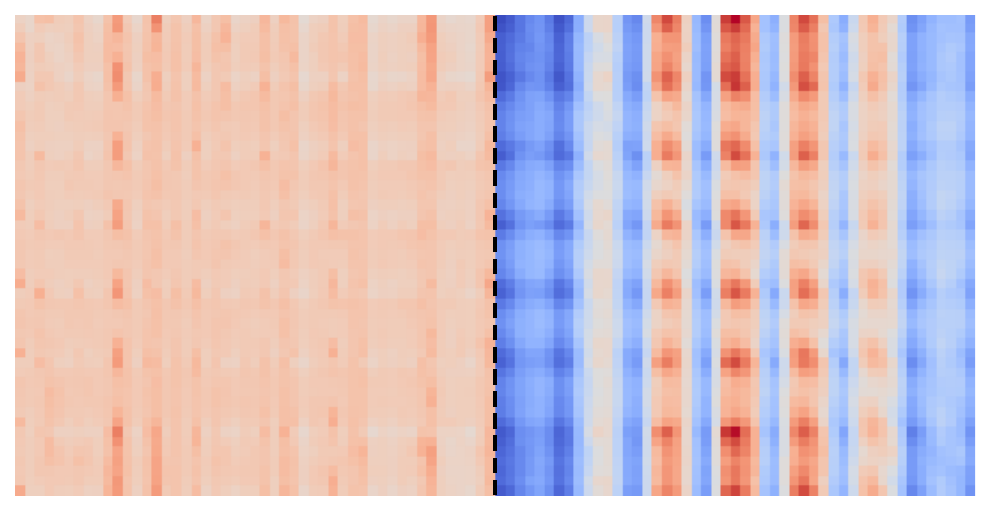}
      \subcaption{Task 4 Epoch 80}
    \end{minipage}
    \hfill
    \begin{minipage}{0.32\textwidth}
      \includegraphics[width=\textwidth]{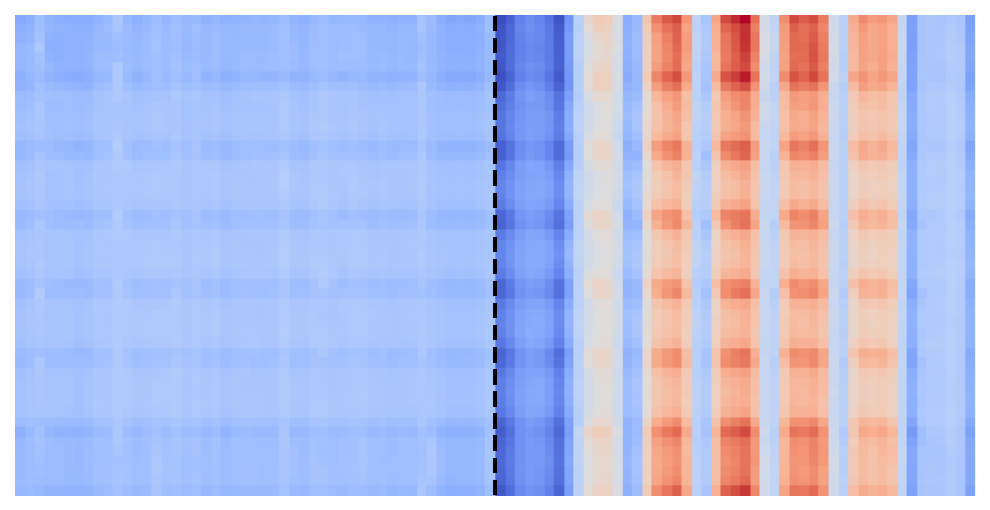}
      \subcaption{Task 4 Epoch 160}
    \end{minipage}
  \end{minipage}
    \caption{Evolution of the attention map averaged across the samples in the merge attention.
    In each sub-figure, each row represents the attention weights for a query patch attending to key patches (red pixels means high attention weights).
    The left part of each sub-figure is the attention weights for task-agnostic key patches ($K_{\mathrm{ta}}$), and the right part is the attention weights for task-specific key patches ($K_{\mathrm{ts}}$), which are separated by a dashed line in the middle.
  From the evolution of the attention maps, we can conclude that
  for each task, \textbf{the high attention values are first on the task-agnostic side, and then transfer to the task-specific side, which represents task-agnostic features are useful for the task at early epochs, and such information is absorbed by the task-specific model at the end of the task.}
  Note that the attention map is averaged across the samples, so the attention map appears more uniform.}
  \label{fig:evo_attn_map}
  \vspace{-2ex}
\end{figure*}

\subsection{Performance Results}
\label{sec:performance}
We test our proposed TagFex and TagFex-P in various class-incremental scenarios.
And we report the accuracy of the test dataset at the end of the entire incremental training (\emph{Last}), and the average test accuracy at the end of each task training across all of the incremental stages (\emph{Avg}).
We run the DER and TagFex experiments for 5 times and report their mean values.
The numerical results are shown in Table~\ref{tab:performance}.
As we can see from the table, TagFex achieves the best results on all of the scenarios, and generally achieves 3$\sim$4\% percent performance gap compared to DER.
The pruned TagFex-P shows minor negative impact on the performance, which effectively reduces the size of the model.
As for the number of parameters, since TagFex does not include extra modules for inference, it holds the same number of parameters as DER.
TagFex-P effectively reduces the averaged incremental number of parameters from 61.6M to 11.6M on CIFAR100 10-10, and from 61.6M to 14.4M on ImageNet100 10-10.
Additionally, the accuracy curves for baselines and TagFex on various scenarios are shown in Figure~\ref{fig:acc_curve}, indicating that TagFex achieve a large performance gap compared to baseline methods.
More comparisons for the number of parameters are shown in Table~\ref{tab:num_params} in Appendix~\ref{app:num_params}.


\subsection{Memory Efficiency}
In this section, we investigate the memory efficiency of our framework.
TagFex saves an extra model for the next task to continually capture the task-agnostic features compared to DER~\cite{yan2021dynamically}.
For ResNet18 on ImageNet, a model occupies the same memory as around 300 rehearsal samples~\cite{zhou2023model}.
It is unfair for methods with less memory saving for the next task.
To eliminate such unfairness, we perform experiments in the memory-aligned protocol, where we compare TagFex to DER with more rehearsal samples.
The results are shown in Table~\ref{tab:memory_aligned_imagenet}.
As we can see from the tables, TagFex is still competitive against DER with comparable number of extra rehearsal samples.
This is more obvious when the memory size is larger (Memory Size 3000, 3300), since TagFex uses the rehearsal samples more effectively than DER.  

\begin{table*}[t]
  \vspace{-3ex}
    \begin{minipage}{0.6\textwidth}
      \centering
\centering
\captionof{table}{Results for TagFex with other self-supervised methods.}
\label{tab:more_ssl}
\scalebox{1.}{
\begin{tblr}{
    row{1} = {c},
    row{2} = {c},
    column{1,3} = {rightsep=15pt},
    cell{1}{1} = {r=2}{},
    cell{1}{2} = {c=2}{},
    cell{1}{4} = {c=2}{},
    cell{3}{2} = {r},
    cell{3}{3} = {r},
    cell{3}{4} = {r},
    cell{3}{5} = {r},
    cell{4}{2} = {r},
    cell{4}{3} = {r},
    cell{4}{4} = {r},
    cell{4}{5} = {r},
    cell{5}{2} = {r},
    cell{5}{3} = {r},
    cell{5}{4} = {r},
    cell{5}{5} = {r},
    cell{6}{2} = {r},
    cell{6}{3} = {r},
    cell{6}{4} = {r},
    cell{6}{5} = {r},
    hline{1,7} = {-}{0.08em},
    hline{2} = {2-5}{},
    hline{3} = {-}{},
}
methods                & 10-10 &       & 50-10 &       \\
                        & Last  & Avg   & Last  & Avg   \\
TagFex                 & 68.23 & 78.45 & 70.33 & 75.87 \\
TagFex w/ VICReg       & 67.85 & 78.21 & 69.68 & 75.44 \\
TagFex w/ Barlow Twins & 68.62 & 78.55 & 70.24 & 75.82 \\
TagFex w/ BYOL         & 69.04 & 78.76 & 71.10 & 76.31 
\end{tblr}}
    \end{minipage}
    \hfill
    \begin{minipage}{0.4\textwidth}
      \centering
      \includegraphics[width=\textwidth]{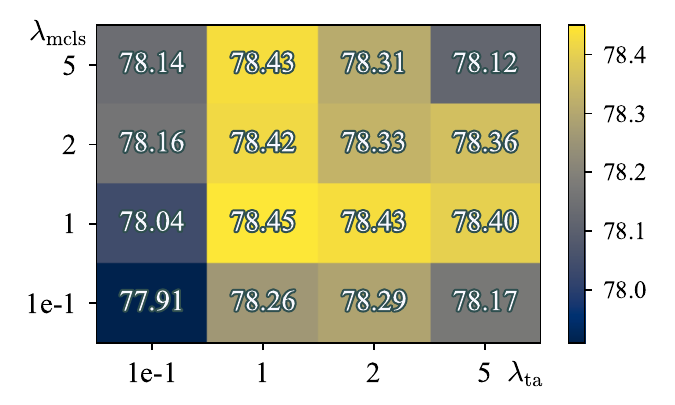}
      \vspace{-4ex}
      \captionof{figure}{Hyperparameter sensitivity analysis on CIFAR100 10-10.}
      \label{fig:hp}
    \end{minipage}
    \vspace{-1ex}
\end{table*}

\begin{table*}[t]
    \begin{minipage}{0.67\textwidth}
        \vspace{2.5ex}
        \centering
    \caption{Ablation study on CIFAR100.}
    \label{tab:ablation}
    \scalebox{0.865}{
    \begin{tblr}{
        row{2} = {c},
        column{1,2,3} = {c},
        cell{1}{1} = {r=2}{},
        cell{1}{2} = {r=2}{},
        cell{1}{3} = {r=2}{},
        cell{1}{4} = {c=2}{c},
        cell{1}{6} = {c=2}{c},
        hline{1,6} = {-}{0.08em},
        hline{2} = {4-7}{},
        hline{3} = {-}{},
    }
    {Task-agnostic Model\\w/ Merge Attention} & {Continual Task-\\agnostic Model} & {Knowledge\\Transferring} & 10-10 &     & 50-10 &    \\
                                &                                                &                            & Last  & Avg   & Last  & Avg   \\
        $\checkmark$         &                $\checkmark$                    &                            & 64.45 & 75.34 & 65.45 & 72.58 \\
        $\checkmark$         &                                                &        $\checkmark$        & 65.86 & 76.32 & 67.51 & 73.29 \\
        $\checkmark$         &                $\checkmark$                    &        $\checkmark$        & 68.23 & 78.45 & 70.33 & 75.87    
    \end{tblr}}
    \end{minipage}
    \hfill
    \begin{minipage}{0.33\textwidth}
        \centering
    \captionof{table}{Memory-aligned average accuracy comparison on ImageNet100.}
    \label{tab:memory_aligned_imagenet}
    \scalebox{0.82}{
      \begin{tblr}{
      rows = {abovesep=1pt, belowsep=1pt},
      row{1} = {c},
      row{2} = {c},
      column{2} = {c},
      cell{1}{1} = {r=2}{},
      cell{1}{2} = {r=2}{},
      cell{1}{3} = {c=2}{},
      hline{1,7} = {-}{0.08em},
      hline{2} = {3-4}{},
      hline{3,5} = {-}{},
    }
    Methods & Memory Size & Scenarios &       \\
            &             & 10-10     & 50-10 \\
    DER     & 2300        & 77.56     & 78.21 \\
    TagFex  & 2000        & \textbf{79.27}     & \textbf{80.64} \\
    DER     & 3300        & 78.43     & 79.12 \\
    TagFex  & 3000        & \textbf{80.87}     & \textbf{81.96} 
    \end{tblr}}
    \end{minipage}
    \vspace{-3ex}
\end{table*}

\subsection{Visualizations on Merge Attention Maps}
\label{sec:merge_attn_vis}
To verify the efficacy of the merge attention, we investigate the evolution of the attention map in the merge attention during the training on ImageNet100.
We record the attention map of some samples at the end of some checkpoint epochs within each task, and see how the attention map evolves during the training of the task.
The results are shown in Figure~\ref{fig:evo_attn_map}.
From the evolution of the attention maps, we can conclude that 
for the incremental tasks the high attention values are first on the task-agnostic side, and then transfer to the task-specific side, which represents task-agnostic features are useful for the task at early epochs, and such information is absorbed by the task-specific model at the end of the task.
The visualizations of more tasks are shown in Appendix~\ref{app:more_vis}.

\subsection{Further Analysis}
\label{sec:further_analysis}
\textbf{TagFex with different strategies for continual learning of the task-agnostic model.}
In section~\ref{sec:acq_ta}, we use SimCLR~\cite{chen2020simple} as the default configuration in the learning of the task-agnostic model.
However, we can use different self-supervised losses on the task-agnostic model.
We test TagFex with three different self-supervised methods on CIFAR100, which is shown in Table~\ref{tab:more_ssl}.
The three methods are VICReg~\cite{bardes2022vicreg} which uses a mix of variance, invariance and covariance regularizations, BYOL~\cite{grill2020bootstrap} which uses a momentum encoder, and Barlow Twins~\cite{zbontar2021barlow} which considers an objective function measuring the cross-correlation matrix between the features.
The results show that TagFex is compatible to different self-supervised methods and performs well with them.
It is worth to note that BYOL and Barlow Twins get better results than the default configuration of TagFex.
However, they are more complicated which require another momentum updated model for training.


\textbf{Hyperparameter analysis}.
TagFex has two hyperparameters $\lambda_{\mathrm{ta}}$ and $\lambda_{\mathrm{mcls}}$ controlling the loss from the task-agnostic model and the merge classifier respectively.
We find the best combination of them through grid search on the average incremental accuracy (Avg).
We perform grid search on CIFAR100 10-10 scenario.
The results are shown in Figure~\ref{fig:hp}.
As we can see from the figure, TagFex is quite robust to these two hyperparameters.
The performance is generally better when $\lambda_{\mathrm{ta}} \ge 1$
The best performance is achieved when their values are balanced and their values are both 1.

\textbf{Ablation study}.
To verify the effectiveness of each part of TagFex, we perform ablation study on three components of TagFex.
Since training the task-agnostic model alone would not influence the task-specific models which are used for inference, and the merge attention cannot work alone, we have to bind them at first (task-agnostic model w/ merge attention).
Then, we add continual learning to the task-agnostic model (continual task-agnostic model), and do not transfer the knowledge from the task-agnostic model to the task-specific model.
Finally, we compare knowledge transferring with and without the continual learning of the task-agnostic model.
We evaluate the configurations on CIFAR100, and compare their average incremental accuracies (Avg).
The results are shown in Table~\ref{tab:ablation}.
As we can see, without the final knowledge transferring, we cannot effectively transfer the information into the task-specific model for inference.
Also, the best performance is achieved with continual learning of task-agnostic model.
\section{Conclusion}
\label{sec:conclusion}
In this paper, we explored the solutions for the feature collision in expansion-based class-incremental learning.
We illustrates the feature collision and propose a framework called TagFex to transfer diverse features into new task-specific model.
In order to acquire such diverse features, we propose to continually train a separate model to capture task-agnostic features in advance for later tasks.
Then, we design a merge attention to adaptively select and aggregate the features to transfer.
Finally, we transfer such features back into the task-specific model.
Furthermore, we not only apply pruning techniques to reduce the number of parameters, but also perform extensive experiments to show the superiority of our proposed framework.
Also, TagFex has competitive memory efficiency against simply adding more rehearsal samples.
From the visualization on merge attention maps, we conclude that the high attention values are transferred from the task-agnostic part to the task-specific part, indicating that the knowledge is absorbed by the task-specific model.
We also perform further analysis to thoroughly evaluate our framework.
\newline
\textbf{Limitations}: Creating a new model for a new task would dramatically increase the number of parameters, even though we could address it by pruning techniques.
In the future work, we would like to seek for more memory efficient methods and techniques.

\section*{Acknowledgements}
{This work is partially supported by National Science and Technology Major Project (2022ZD0114805), NSFC (62476123, 62376118, 62006112, 62250069), Fundamental Research Funds for the Central Universities (2024300373,14380021), CCF-Tencent Rhino-Bird Open Research Fund RAGR20240101, Collaborative Innovation Center of Novel Software Technology and Industrialization.}

\nocite{zhang2024fscil, zhang2024deep, wang2025ipfa}

{
    \small
    \bibliographystyle{ieeenat_fullname}
    \bibliography{main}
}

\clearpage
\setcounter{page}{1}
\maketitlesupplementary

\section{Additional Related Works}
\label{app:rel_works}

In Section~\ref{sec:related_works}, we discuss related works closely related to this work.
Here we provide some additional related works in CIL.
\textbf{Rehearsal memory} is used to store exemplars of previous tasks and replay at follow-up tasks.
It makes the learned feature less forgetful by adjusting the input distribution towards the learned tasks.
Many works focus on how to select exemplars~\cite{rebuffi2017icarl, wu2019large, tiwari2022gcr,liu2020mnemonics}. 
Exemplars can also be obtained by generative models~\cite{shin2017continual}.

\textbf{Regularization} methods~\cite{kirkpatrick2017overcoming,zhu2021class,shi2022mimicking} come from various ideas.
\cite{kirkpatrick2017overcoming} proposes to restrict the updates of important parameters.
\cite{zhu2021class} proposes a dual augmentation framework to make the eigenvalues of the representation's covariance matrix larger.
\cite{shi2022mimicking} proposes to make the representation scatter uniformly, making the representation contains more information about the input sample. 

\textbf{Model distillation} uses the model trained on previous tasks as a teacher and distillation losses to keep the previously learned knowledge in the feature.
LwF~\cite{li2017learning} proposes to use the response of the old model to guide the training of the new model's old tasks. 
PODNet~\cite{douillard2020podnet} uses the pooled intermediate feature maps of the ResNet to be the distillation target in training.

\textbf{Pre-Trained Model-based} methods leverage pretrained models and adapt the model for class-incremental learning~\cite{wang2022learning, wang2022sprompts, wang2022dualprompt}. 
\cite{wang2022learning} uses visual prompt tuning~\cite{jia2022visual} to learn a prompt for each task.
\cite{wang2022dualprompt} proposes the dual prompt scheme in ViT.
Due to the head start of the pre-trained models in learning representations, these methods outperform the methods which train the model from scratch, even without the rehearsal memory samples.

Other perspectives to boost CIL are also considered.
In the parameter space, \cite{mirzadeh2020linear} studies the linear mode connectivity in CIL and proposes to enhance the linear mode connectivity between learned models.
\cite{lin2022towards} also considers the linear mode connectivity between learned models and proposes to combine two models learned in different ways to get better linear mode connectivity.

\cite{dhar2019learning} uses Grad-CAM to generate attention maps for distillation.
\cite{goswami2024resurrecting} generates adversarial samples iteratively to perform drift estimation for old class prototypes.
\cite{rymarczyk2023icicle} leverages the part information of the images, forming prototypical part layers on the model, improving the interpretability of the model.
\cite{zheng2023preserving} proposes a locality-preserving attention module to remedy the locality degradation during the training of CIL.
\cite{zheng2024multi} considers enlarging the all-layer margin on the rehearsal samples, applying feature augmentations by input gradients.

\section{More Implementation Details}
\label{app:more_imp}
In Section~\ref{sec:exp_setups}, we describe the experiment settings for the experiment section.
Here we provide more implementation details.
We provide more configurations about TagFex in Table~\ref{tab:imp_detail}.
We use the same configurations for CIFAR and ImageNet unless additional specification.
We use AutoAugment~\cite{cubuk2019autoaugment} for creating augmented samples in the training of the task-agnostic model.
We perform our experiments on a server with 4 RTX3090 GPUs.
For the pruning threshold, we adaptively choose the threshold according to the expected compression rate 0.4.

\section{Performance Results with Standard Deviations.}
\label{app:std_dev}
In Section~\ref{sec:performance}, we compare the performance of TagFex with some baselines.
Here, we provide performance results on CIFAR100 and ImageNet100 with standard deviations of 5 runs in Table~\ref{tab:std_dev}.

\section{Comparison for the Number of Parameters}
\label{app:num_params}
In Section~\ref{sec:performance} and Table~\ref{tab:performance}, we provide the performance results of pruned TagFex-P on various scenarios, and describe the efficacy on reducing the number of parameters.
Here, we provide detailed comparison for the average number of parameters of the inference model.
It is shown in Table~\ref{tab:num_params}.
We compare the average incremental number of parameters for inference, which the values are averaged across each stages.
As we can see, TagFex holds the same number of parameters as DER.
With pruning techniques, TagFex-P requires less number of parameters to achieve comparable performance.
\begin{table}[h]
    \centering
    \caption{Comparison for number of parameters. Values are shown in millions.}
    \label{tab:num_params}
    \begin{tblr}{
        row{1} = {c},
        row{2} = {c},
        column{3} = {rightsep=15pt},
        cell{1}{1} = {r=2}{},
        cell{1}{2} = {c=2}{},
        cell{1}{4} = {c=2}{},
        cell{3}{2} = {c},
        cell{3}{3} = {c},
        cell{3}{4} = {c},
        cell{3}{5} = {c},
        cell{4}{2} = {c},
        cell{4}{3} = {c},
        cell{4}{4} = {c},
        cell{4}{5} = {c},
        cell{5}{2} = {c},
        cell{5}{3} = {c},
        cell{5}{4} = {c},
        cell{5}{5} = {c},
        cell{6}{2} = {c},
        cell{6}{3} = {c},
        cell{6}{4} = {c},
        cell{6}{5} = {c},
        cell{7}{2} = {c},
        cell{7}{3} = {c},
        cell{7}{4} = {c},
        cell{7}{5} = {c},
        cell{8}{2} = {c},
        cell{8}{3} = {c},
        cell{8}{4} = {c},
        cell{8}{5} = {c},
        hline{1,9} = {-}{0.08em},
        hline{2} = {2-5}{},
        hline{3,7} = {-}{},
    }
    Methods                         & CIFAR100 &       & ImageNet100 &       \\
                                    & 10-10    & 50-10 & 10-10       & 50-10 \\
    iCaRL~\cite{rebuffi2017icarl}   & 0.46     & 0.46  & 11.2        & 11.2  \\
    BiC~\cite{wu2019large}          & 0.46     & 0.46  & 11.2        & 11.2  \\
    DyTox~\cite{douillard2022dytox} & 10.7     & -     & 11.0        & -     \\
    DER~\cite{yan2021dynamically}   & 61.6     & 39.2  & 61.6        & 39.2  \\
    TagFex                          & 61.6     & 39.2  & 61.6        & 39.2  \\
    TagFex-P                        & 11.6     & 9.8  & 14.4        & 11.3  
    \end{tblr}
    \end{table}

\begin{figure*}[t]
  \begin{minipage}{0.245\textwidth}
    \includegraphics[width=\textwidth]{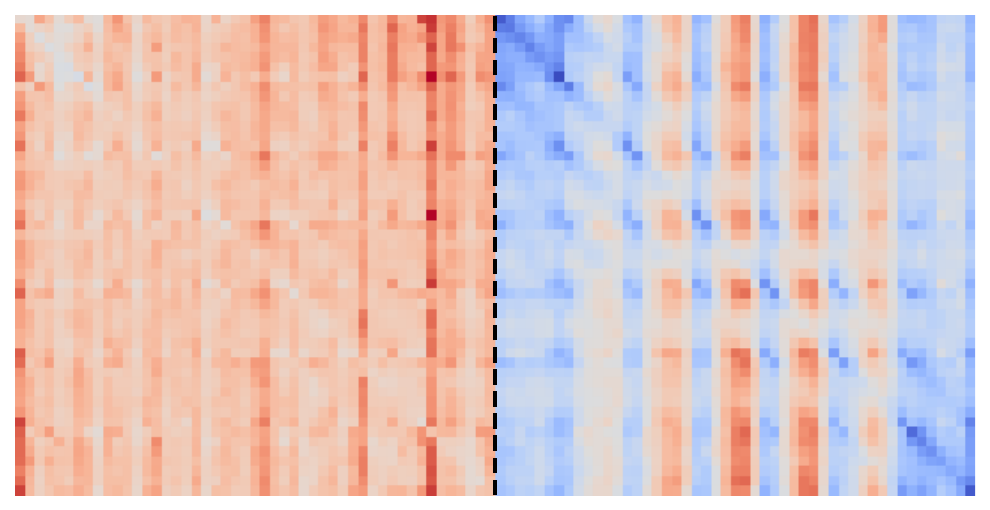}
    \subcaption{Task 1 Epoch 0}
  \end{minipage}
  \hfill
  \begin{minipage}{0.245\textwidth}
    \includegraphics[width=\textwidth]{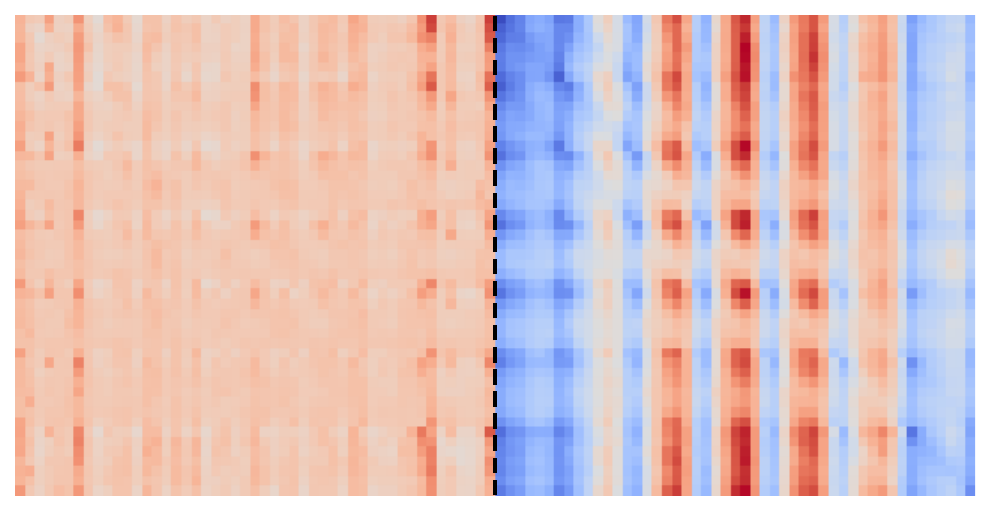}
    \subcaption{Task 1 Epoch 60}
  \end{minipage}
  \hfill
  \begin{minipage}{0.245\textwidth}
    \includegraphics[width=\textwidth]{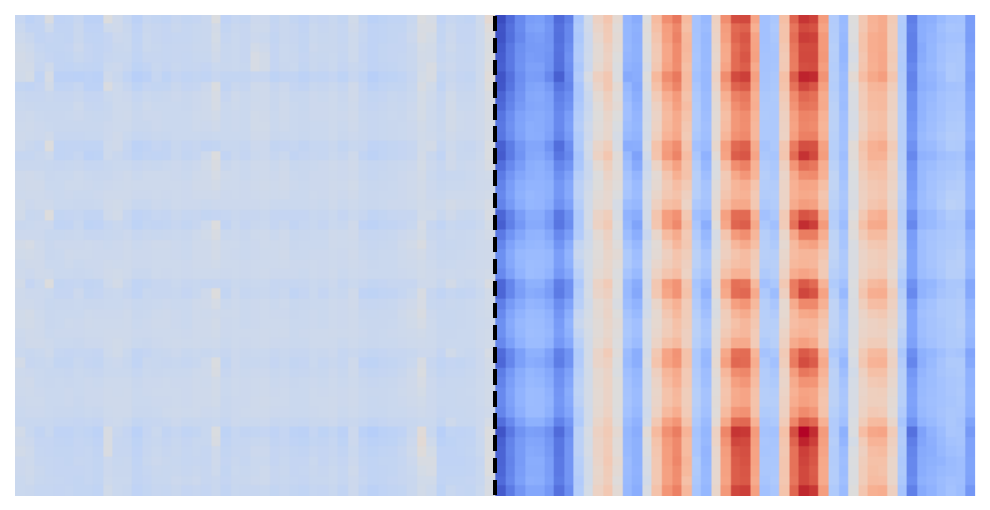}
    \subcaption{Task 1 Epoch 120}
  \end{minipage}
  \hfill
  \begin{minipage}{0.245\textwidth}
    \includegraphics[width=\textwidth]{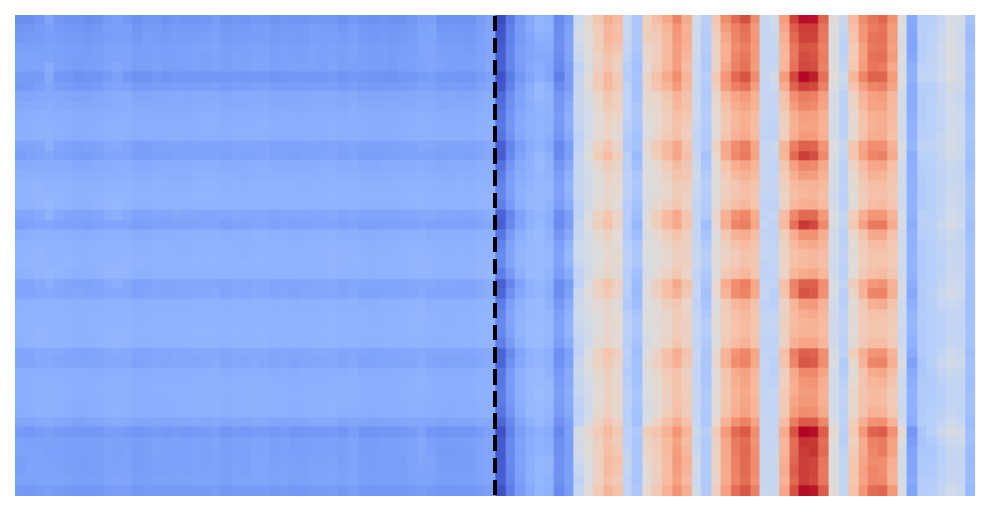}
    \subcaption{Task 1 Epoch 160}
  \end{minipage}
  \newline
  \begin{minipage}{0.245\textwidth}
    \includegraphics[width=\textwidth]{./figures/merge_attn/attn_map_evo_task2_ep0.pdf}
    \subcaption{Task 2 Epoch 0}
  \end{minipage}
  \hfill
  \begin{minipage}{0.245\textwidth}
    \includegraphics[width=\textwidth]{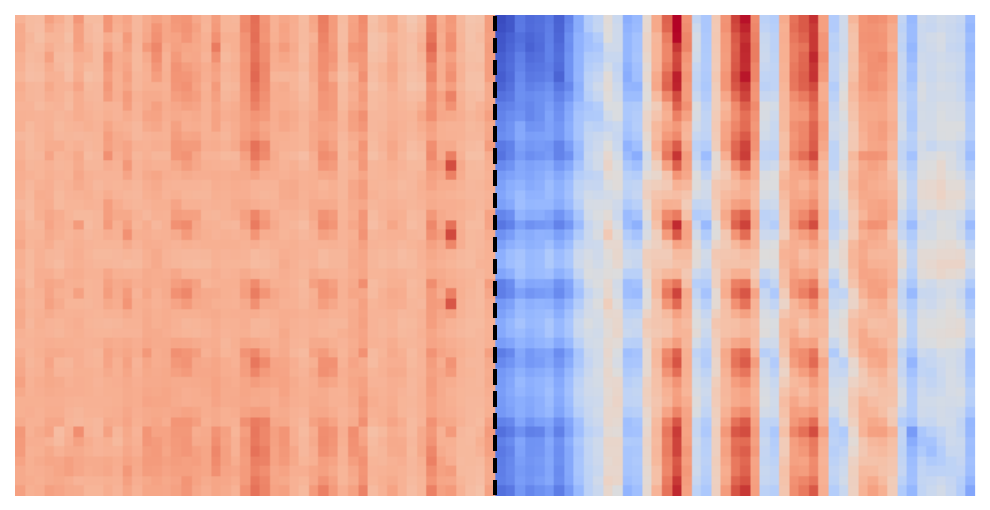}
    \subcaption{Task 2 Epoch 60}
  \end{minipage}
  \hfill
  \begin{minipage}{0.245\textwidth}
    \includegraphics[width=\textwidth]{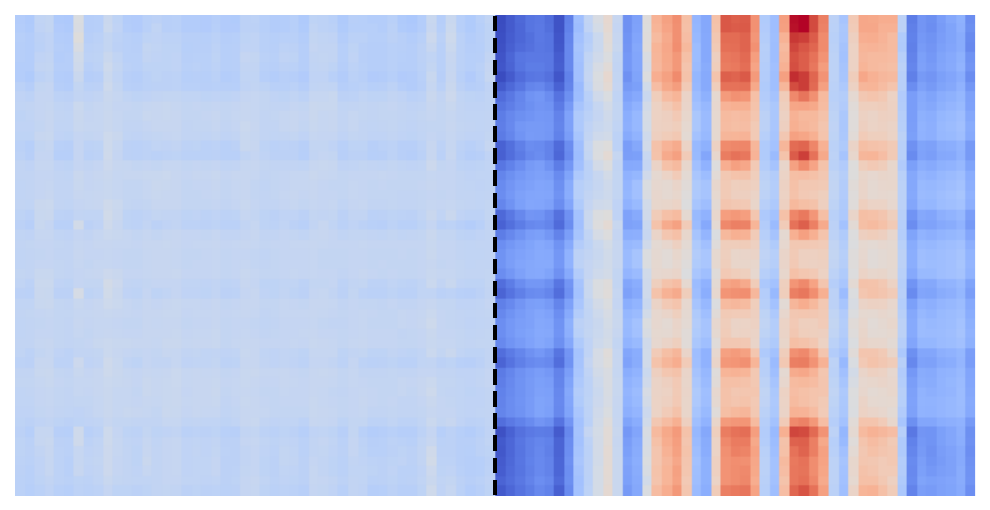}
    \subcaption{Task 2 Epoch 120}
  \end{minipage}
  \hfill
  \begin{minipage}{0.245\textwidth}
    \includegraphics[width=\textwidth]{./figures/merge_attn/attn_map_evo_task2_ep160.pdf}
    \subcaption{Task 2 Epoch 160}
  \end{minipage}
  \newline
  \begin{minipage}{0.245\textwidth}
    \includegraphics[width=\textwidth]{./figures/merge_attn/attn_map_evo_task4_ep0.pdf}
    \subcaption{Task 4 Epoch 0}
  \end{minipage}
  \hfill
  \begin{minipage}{0.245\textwidth}
    \includegraphics[width=\textwidth]{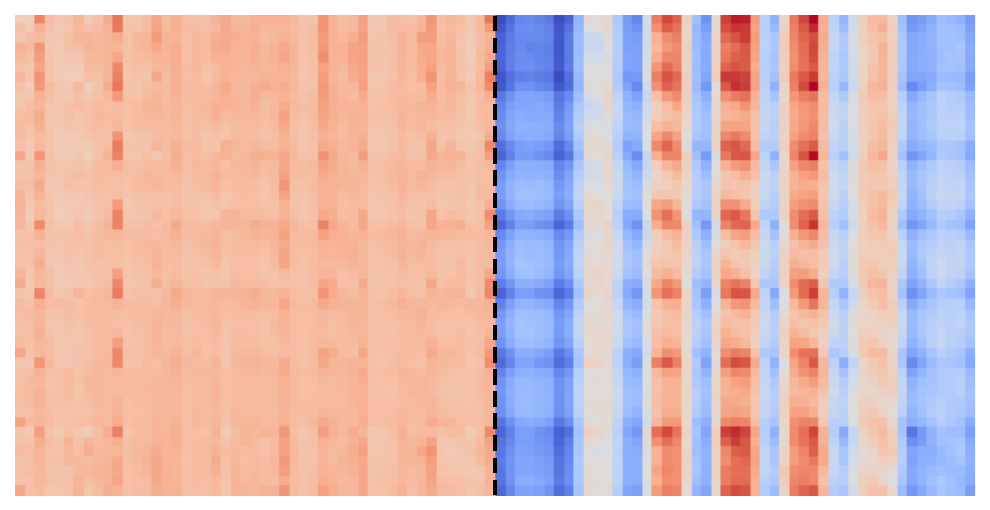}
    \subcaption{Task 4 Epoch 60}
  \end{minipage}
  \hfill
  \begin{minipage}{0.245\textwidth}
    \includegraphics[width=\textwidth]{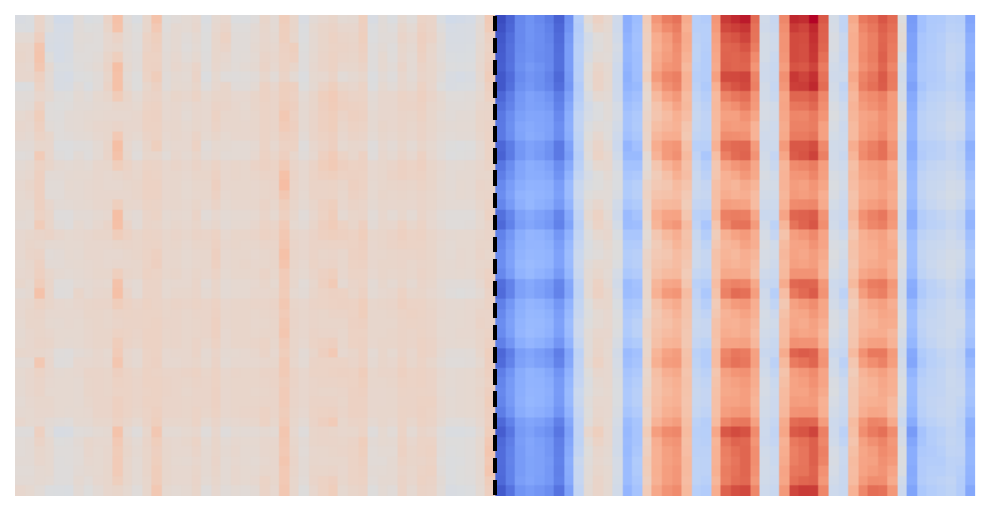}
    \subcaption{Task 4 Epoch 120}
  \end{minipage}
  \hfill
  \begin{minipage}{0.245\textwidth}
    \includegraphics[width=\textwidth]{./figures/merge_attn/attn_map_evo_task4_ep160.pdf}
    \subcaption{Task 4 Epoch 160}
  \end{minipage}
  \newline
  \begin{minipage}{0.245\textwidth}
    \includegraphics[width=\textwidth]{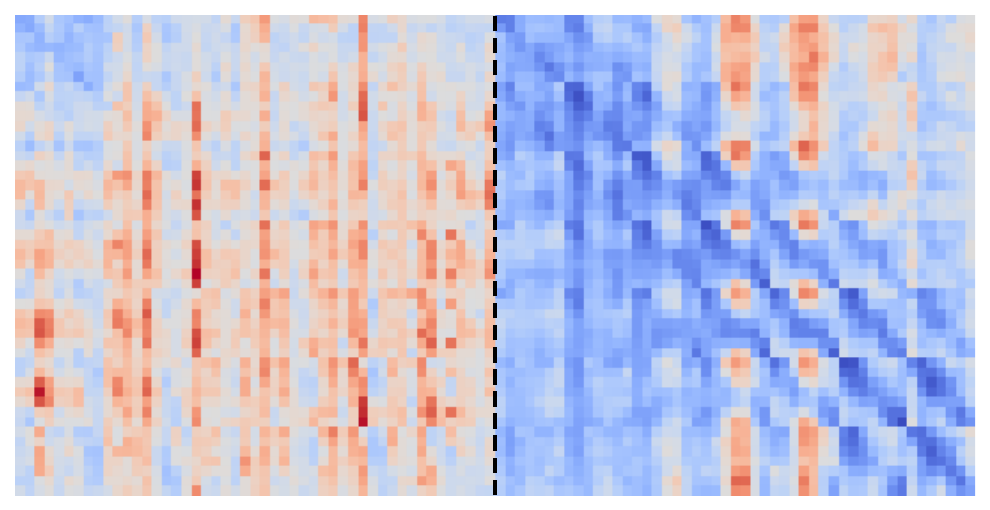}
    \subcaption{Task 6 Epoch 0}
  \end{minipage}
  \hfill
  \begin{minipage}{0.245\textwidth}
    \includegraphics[width=\textwidth]{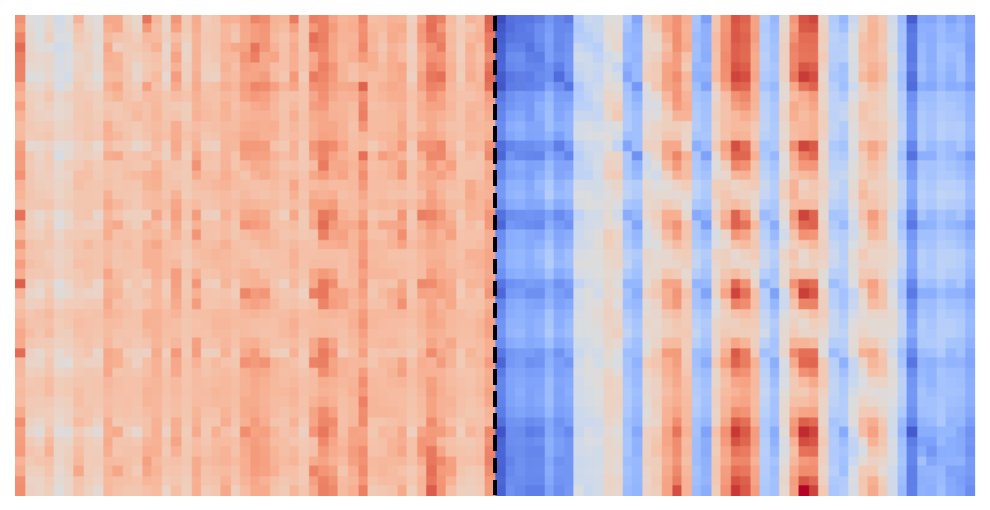}
    \subcaption{Task 6 Epoch 60}
  \end{minipage}
  \hfill
  \begin{minipage}{0.245\textwidth}
    \includegraphics[width=\textwidth]{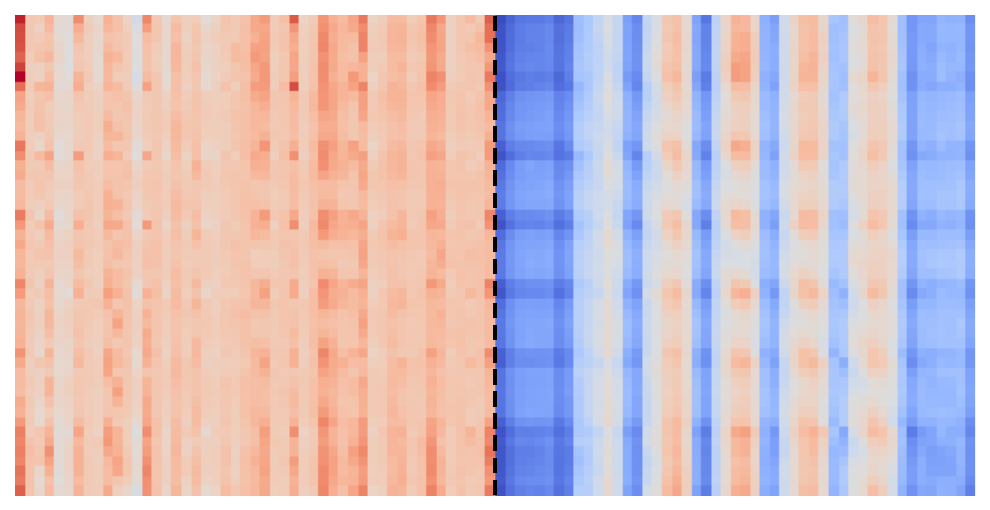}
    \subcaption{Task 6 Epoch 120}
  \end{minipage}
  \hfill
  \begin{minipage}{0.245\textwidth}
    \includegraphics[width=\textwidth]{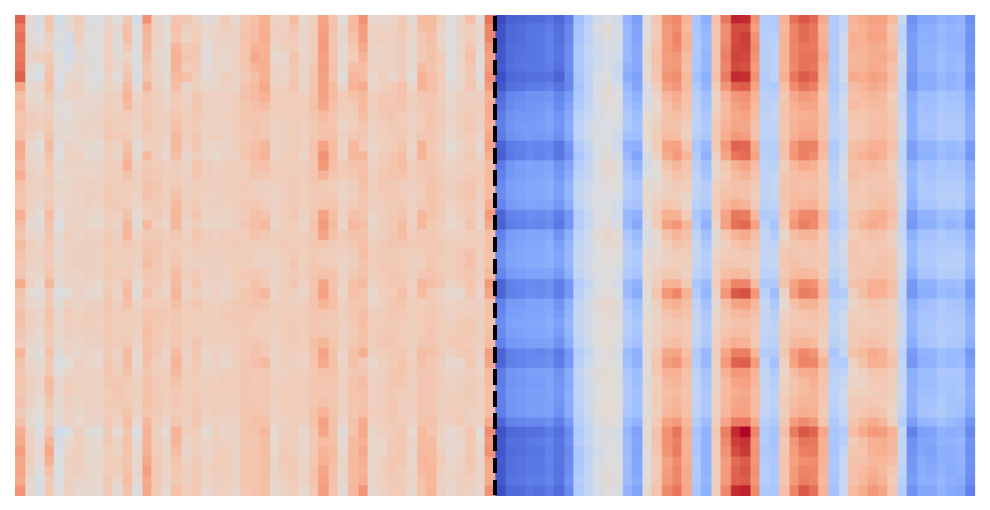}
    \subcaption{Task 6 Epoch 160}
  \end{minipage}
  \caption{More visualizations on merge attention maps.}
  \vspace{-3ex}
  \label{fig:more_attn_map}
\end{figure*}

\section{CKA Similarities of Learned Features in Expansion-based Methods}
\label{app:cka}
\begin{figure*}[t]
  \begin{minipage}{0.495\textwidth}
    \includegraphics[width=\textwidth]{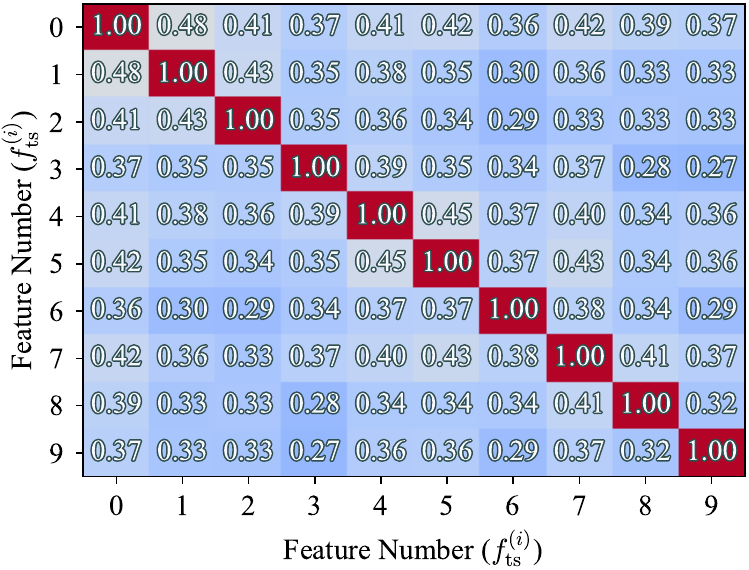}
    \subcaption{CKA feature similarities between each model learned by DER.}
  \end{minipage}
  \begin{minipage}{0.495\textwidth}
    \includegraphics[width=\textwidth]{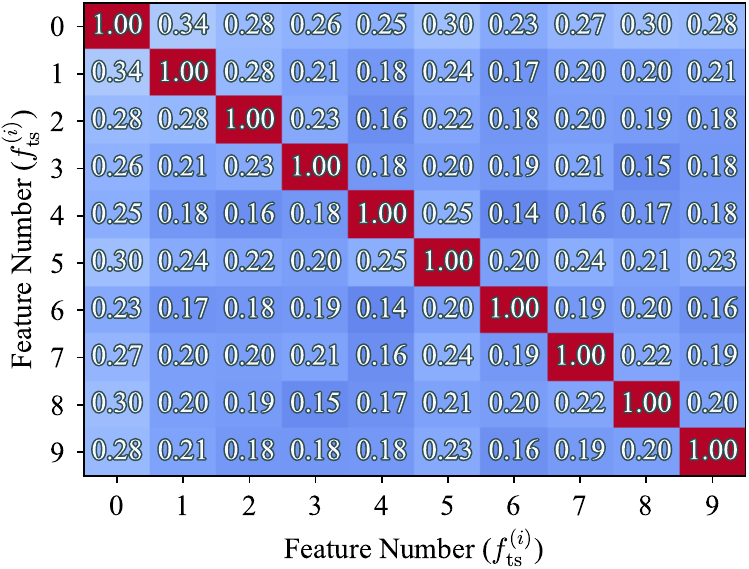}
    \subcaption{CKA feature similarities between each model learned by TagFex.}
  \end{minipage}
  \caption{CKA feature similarities. Features learned by TagFex are more diverse and less correlated.}
  \label{fig:cka_number}
\end{figure*}

In Section~\ref{sec:introduction}, we compare the average CKA similarities of the learned features in expansion-based methods.
Here we show the detailed numbers for the CKA similarities between each expanded feature.
The results are shown in Figure~\ref{fig:cka_number}, from which we can conclude that the features extracted by the models learned by DER are relatively similar, with maximum CKA 0.48 and minimum 0.27, especially when comparing to TagFex, with maximum 0.34 and minimum 0.14.

\section{More Visualizations on Merge Attention Maps}
\label{app:more_vis}
In Section~\ref{sec:merge_attn_vis}, we discuss the evolution of the merge attention maps for TagFex.
Here we provide more visualizations on merge attention maps in Figure~\ref{fig:more_attn_map}.
As we can see from the figures, in each incremental task, the high attention values are first on the task-agnostic side, and then transfer to the task-specific side, which represents task-agnostic features are useful for the task at early epochs, and such information is absorbed by the task-specific model.
In addition, for later tasks (task 6 in the figure, sub-figure (m)(n)(o)(p)), the attention on the task-agnostic side at the end of the task is not as small as previous tasks (sub-figure (d)(h)(l)).
It indicates that the task-agnostic model contains more and more features during the continual self-supervised learning.

\section{Performance Experiments on Fine-grained Dataset}

To verify the performance gain of TagFex on fine-grained datasets, we perform experiments on CUB200 100-20 with memory size 2000.
The results are shown in Table~\ref{tab:cub200}, which shows strong performance gain over DER.

\begin{table}[h]
  \caption{Performance results on CUB200. TagFex outperforms DER on such fine-grained dataset.}
  \label{tab:cub200}
  \centering
  \scalebox{1.0}{
  \begin{tblr}{
      row{2} = {c},
      cell{1}{1} = {r=2}{},
      cell{1}{2} = {c=2}{c},
      cell{3}{2} = {c},
      cell{3}{3} = {c},
      cell{4}{2} = {c},
      cell{4}{3} = {c},
      hline{1,5} = {-}{0.08em},
      hline{2} = {2-3}{},
      hline{3} = {-}{},
      }
      Methods & CUB200 100-20 &       \\
              & Avg           & Last  \\
      DER    & 53.07         & 52.88 \\
      TagFex & 56.56         & 54.37 
      \end{tblr}}
\end{table}

\begin{table*}[t]
    \centering
    \caption{More implementation details for TagFex.}
    \label{tab:imp_detail}
    \begin{tblr}{
      row{1} = {c},
      column{2} = {rightsep=20pt},
      hline{1,7} = {-}{0.08em},
      hline{2,6} = {-}{},
    }
    General Configurations      &      & Task-Agnostic Model          &                 \\
    Base Epochs                 & 200  & InfoNCE Temperature $\tau$ & 0.2             \\
     Incremental~Epochs         &  170 & Projection Hidden Dim.       & 2048     \\
    Batch Size                  & 128  & Projection Embedding Dim.    & 1024     \\
    Learning Rate               & 0.1  & Predictor       & Linear     \\
    Merge Attention \# of heads & 8    & Transferring Temperature     & 2               
    \end{tblr}
    \end{table*}
\begin{table*}[t]
    \centering
    \caption{Performance results on CIFAR100 and ImageNet100 with standard deviations.}
    \label{tab:std_dev}
    \begin{tblr}{
      row{2} = {c},
      column{2,4} = {rightsep=20pt},
      cell{1}{1} = {r=2}{},
      cell{1}{2} = {r=2}{},
      cell{1}{3} = {c=2}{c},
      cell{1}{5} = {c=2}{c},
      cell{3}{1} = {r=3}{},
      cell{3}{3} = {c},
      cell{3}{4} = {c},
      cell{3}{5} = {c},
      cell{3}{6} = {c},
      cell{4}{3} = {c},
      cell{4}{4} = {c},
      cell{4}{5} = {c},
      cell{4}{6} = {c},
      cell{5}{3} = {c},
      cell{5}{4} = {c},
      cell{5}{5} = {c},
      cell{5}{6} = {c},
      cell{6}{1} = {r=3}{},
      cell{6}{3} = {c},
      cell{6}{4} = {c},
      cell{6}{5} = {c},
      cell{6}{6} = {c},
      cell{7}{3} = {c},
      cell{7}{4} = {c},
      cell{7}{5} = {c},
      cell{7}{6} = {c},
      cell{8}{3} = {c},
      cell{8}{4} = {c},
      cell{8}{5} = {c},
      cell{8}{6} = {c},
      hline{1,9} = {-}{0.08em},
      hline{2} = {3-6}{},
      hline{3,6} = {-}{},
    }
    Datasets    & Methods  & 10-10                      &                            & 50-10                      &                            \\
                &          & Last                       & Avg                        & Last                       & Avg                        \\
    CIFAR100    & DER      & 64.35\scriptsize $\pm$0.11 & 75.36\scriptsize $\pm$0.06 & 65.27\scriptsize $\pm$0.10 & 72.60\scriptsize $\pm$0.05 \\
                & TagFex   & 68.23\scriptsize $\pm$0.13 & 78.45\scriptsize $\pm$0.06 & 70.33\scriptsize $\pm$0.13 & 75.87\scriptsize $\pm$0.06 \\
                & TagFex-P & 67.34\scriptsize $\pm$0.18 & 78.02\scriptsize $\pm$0.10 & 69.26\scriptsize $\pm$0.15 & 74.24\scriptsize $\pm$0.08 \\
    ImageNet100 & DER      & 66.71\scriptsize $\pm$0.16 & 77.18\scriptsize $\pm$0.08 & 71.08\scriptsize $\pm$0.11 & 77.71\scriptsize $\pm$0.06 \\
                & TagFex   & 70.84\scriptsize $\pm$0.19 & 79.27\scriptsize $\pm$0.11 & 75.54\scriptsize $\pm$0.11 & 80.64\scriptsize $\pm$0.07 \\
                & TagFex-P & 69.21\scriptsize $\pm$0.17 & 78.56\scriptsize $\pm$0.10 & 74.13\scriptsize $\pm$0.12 & 79.85\scriptsize $\pm$0.07 
    \end{tblr}
    \end{table*}

\end{document}